\documentclass[10pt,twocolumn,letterpaper]{article}

\usepackage[pagenumbers]{cvpr} 
\usepackage{graphicx}
\usepackage{amsmath}
\usepackage{amssymb}
\usepackage{booktabs}
\usepackage{bm}
\usepackage{epsfig}
\makeatletter
\@namedef{ver@everyshi.sty}{}
\makeatother
\usepackage{tikz}
\usetikzlibrary{spy}
\usepackage{algpseudocode}
\usepackage{algorithm}
\usepackage{amsthm}
\usepackage{mathrsfs}
\usepackage{nicefrac}       
\usepackage{booktabs}       
\usepackage{adjustbox}
\usepackage{makecell}
\usepackage{tabularx}
\usepackage{wrapfig}
\usepackage{pgfplots}
\usepackage{thmtools,thm-restate}

\declaretheorem[]{remark}

\usetikzlibrary{positioning}
\usepackage{subcaption}
\usepackage[pagebackref,breaklinks,colorlinks]{hyperref}

\newcommand{\xb}{{\boldsymbol x}}
\newcommand{\yb}{{\boldsymbol y}}

\newcommand{\nb}{{\boldsymbol n}}
\newcommand{\kb}{{\boldsymbol k}}

\newcommand{\s}{{\boldsymbol s}}
\newcommand{\n}{{\boldsymbol n}}
\newcommand{\x}{{\boldsymbol x}}
\newcommand{\y}{{\boldsymbol y}}
\newcommand{\z}{{\boldsymbol z}}
\newcommand{\w}{{\boldsymbol w}}

\newcommand{\Kb}{{\boldsymbol K}}
\newcommand{\Xb}{{\boldsymbol X}}

\newcommand{\Ib}{{\boldsymbol I}}

\newcommand{\mub}{{\bm \mu}}
\newcommand{\phib}{{\bm \phi}}
\newcommand{\varphib}{{\bm \varphi}}

\newcommand{\Ed}{{\mathbb E}}
\newcommand{\Rd}{{\mathbb R}}

\newcommand{\Hc}{{\mathcal H}}
\newcommand{\Nc}{{\mathcal N}}

\newcommand{\Pc}{{\mathcal P}}
\newcommand{\Tc}{{\mathcal T}}
\newcommand{\Lc}{{\mathcal L}}
\newcommand{\Jc}{{\mathcal J}}

\DeclareMathOperator*{\argmin}{\arg\!\min}

\newcommand{\code}[1] {\texttt{#1}}

\definecolor{C0}{rgb}{0.121569, 0.466667, 0.705882}
\definecolor{C1}{rgb}{1.000000, 0.498039, 0.054902}
\definecolor{C2}{rgb}{0.172549, 0.627451, 0.172549}
\definecolor{C3}{rgb}{0.839216, 0.152941, 0.156863}
\definecolor{C4}{rgb}{0.580392, 0.403922, 0.741176}
\definecolor{C5}{rgb}{0.549020, 0.337255, 0.294118}
\definecolor{C6}{rgb}{0.890196, 0.466667, 0.760784}
\definecolor{C7}{rgb}{0.498039, 0.498039, 0.498039}
\definecolor{C8}{rgb}{0.737255, 0.741176, 0.133333}
\definecolor{C9}{rgb}{0.090196, 0.745098, 0.811765}
\definecolor{trolleygrey}{rgb}{0.5, 0.5, 0.5}

\usepackage[capitalize]{cleveref}
\crefname{section}{Sec.}{Secs.}
\Crefname{section}{Section}{Sections}
\Crefname{table}{Table}{Tables}
\crefname{table}{Tab.}{Tabs.}

\def\cvprPaperID{7278} 
\def\confName{CVPR}
\def\confYear{2023}

\begin{document}

\title{Parallel Diffusion Models of Operator and Image for Blind Inverse Problems}

\author{Hyungjin Chung\textsuperscript{\rm 1}\thanks{Equal contribution}{ \ }\quad\quad Jeongsol Kim\textsuperscript{\rm 1}$^{*}$\quad\quad Sehui Kim\textsuperscript{\rm 2} \quad\quad Jong Chul Ye\textsuperscript{\rm 2}\\
\textsuperscript{\rm 1}Dept. of Bio \& Brain Engineering,\\
\textsuperscript{\rm 2}Kim Jae Chul Graduate School of AI,\\
KAIST \\
{\tt\small \{hj.chung, jeongsol, sehui.kim, jong.ye\}@kaist.ac.kr}}
\twocolumn[{%
\vspace{-0.5cm}
\renewcommand\twocolumn[1][]{#1}%
\maketitle
\begin{center}
\vspace{-1.0cm}
\includegraphics[width=1.01\linewidth]{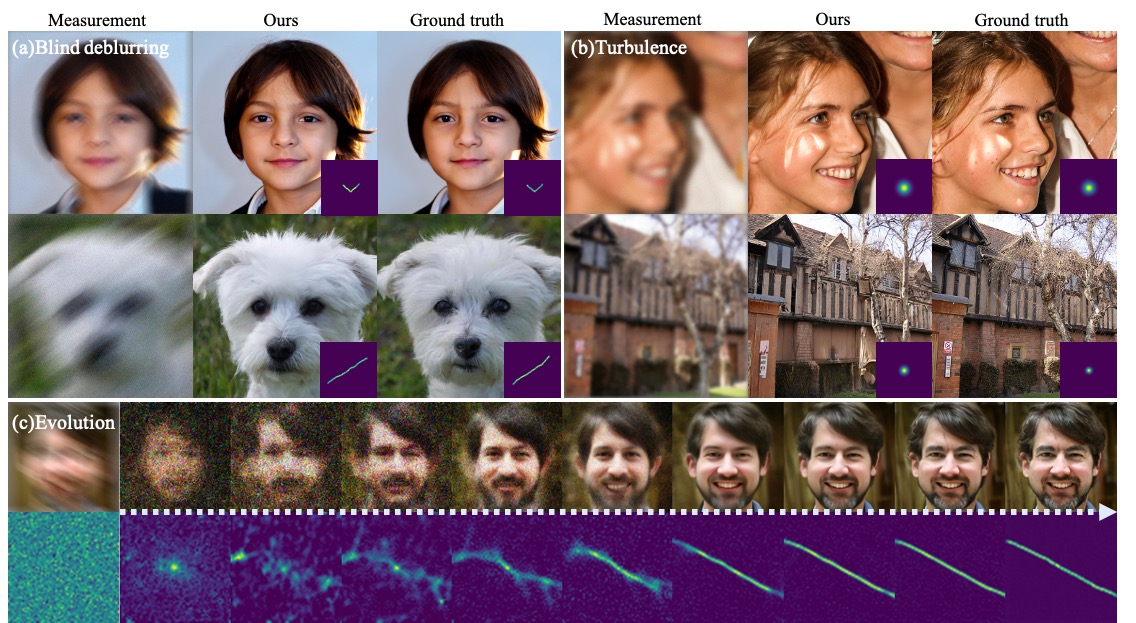}
\captionof{figure}{{Representative results and overall concept of the proposed method. (a) Results of blind deblurring. Both the image and the kernel in the bottom right corner are jointly estimated with the proposed method. (b) Results of imaging through turbulence. (c) Evolution of joint reconstruction with the proposed method. 1$^{\rm st}$, 2$^{\rm nd}$ row illustrate the change of $\hat\x_0(\x_t)$ and $\hat\kb_0(\kb_t)$ through time as $t = 1 \rightarrow 0$, with the measurement and the kernel initialization given on the first column.}}
\label{fig:cover}
\end{center}
}]

\begin{abstract}
Diffusion model-based inverse problem solvers have demonstrated state-of-the-art performance in cases where the forward operator is known (i.e. non-blind). However, the applicability of the method to blind inverse problems has yet to be explored. In this work, we show that we can indeed solve a family of blind inverse problems by constructing another diffusion prior for the forward operator. Specifically, parallel reverse diffusion guided by gradients from the intermediate stages enables joint optimization of both the forward operator parameters as well as the image, such that both are jointly estimated at the end of the parallel reverse diffusion procedure. We show the efficacy of our method on two representative tasks --- blind deblurring, and imaging through turbulence --- and show that our method yields state-of-the-art performance, while also being flexible to be applicable to general blind inverse problems when we know the functional forms.
\end{abstract}

\begin{figure*}[t]
  \centering
    \centerline{{\includegraphics[width=0.8\linewidth]{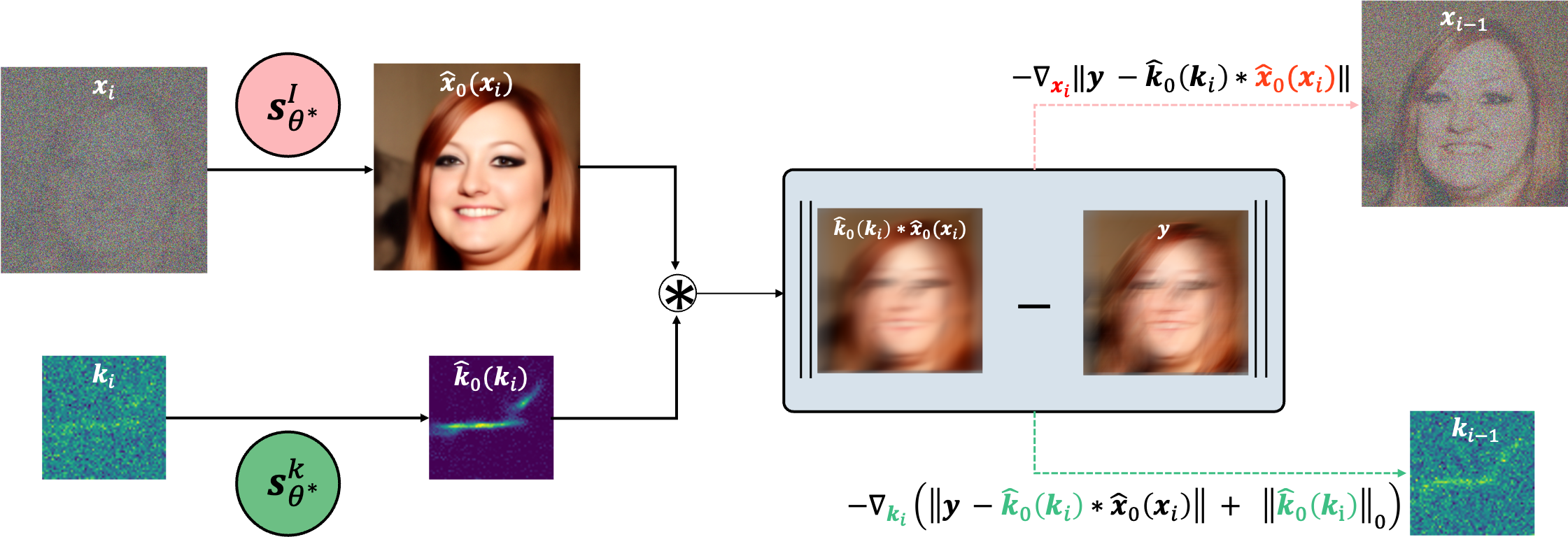}}}
  \caption{{Description of BlindDPS. From the intermediate (noisy) estimate $\x_i, \kb_i$, we achieve the denoised representation $\hat\x_0(\x_i), \hat\kb_0(\kb_i)$ through Tweedie's formula with the score functions $\s_{\theta^*}^i, \s_{\theta^*}^k$. The residual $\|\y - \hat\kb_0 \ast \hat\x_0\|$ is computed with the denoised estimates, and the residual-minimizing gradients are applied parallel to both diffusion processes.}}
  \label{fig:method}
  \vspace{-0.5cm}
\end{figure*}

\section{Introduction}
\label{sec:intro}

Inverse problems subsume a wide set of important problems in science and engineering, where the objective is to recover the latent image from the corrupted measurement, generated by the forward operator.
Considering the taxonomy, they can be split into two major categories --- {\em non-blind} inverse problems, and {\em blind} inverse problems. The former considers the cases where the forward operator is known, and hence eases the problem. In contrast, the latter considers the cases where the operator is {\em unknown}, and thus the operator needs to be estimated together with the reconstruction of the latent image. The latter problem is considerably harder than the former problem, as joint minimization is typically much less stable.

In this work, we mainly focus on leveraging generative priors to solve inverse problems in imaging. Among many different generative model classes, diffusion models have established the new state-of-the-art. In diffusion models, we define the {\em forward} data noising process, which gradually corrupts the image into white Gaussian noise. The generative process is defined by the {\em reverse} of such process, where each step of reverse diffusion is governed by the score function~\cite{song2020score}. 
With the recent surge of diffusion models, it has been demonstrated in literature that diffusion models are not only powerful generative models, but also excellent generative priors to solve inverse problems. Namely, one can either resort to iterative projections to the measurement subspace~\cite{song2020score,chung2022come}, or estimate posterior sampling~\cite{chung2022diffusion} to arrive at feasible solutions that meet the data consistency. For both linear~\cite{song2020score,chung2022come,kawar2022denoising} and some non-linear~\cite{chung2022diffusion,anonymous2023pseudoinverseguided} inverse problems, guiding unconditional diffusion models to solve down-stream inverse problems were shown to have stronger performance even when compared to the fully supervised counterparts.

Nevertheless, current solvers are strictly limited to cases where the forward operator is known and fixed. For example, \cite{kawar2022denoising,chung2022diffusion} consider non-blind deblurring with known kernels. The problem now boils down to optimizing only for the latent image, since the likelihood can be computed robustly. Unfortunately, in real world problems, knowing the kernel exactly is impractical. It is often the case where the kernel is also unknown, and we have to jointly estimate the image and the kernel. In such cases, not only do we need a prior model of the image, but we also need some proper prior model of the kernel~\cite{pan2016l_0, sun2013edge}. While conventional methods exploit, e.g. patch-based prior~\cite{sun2013edge}, sparsity prior~\cite{pan2016l_0}, etc., they often fall short of accurate modeling of the distribution.

In this work, we aim to leverage the ability of diffusion models to act as strong generative priors and propose {\em BlindDPS} (Blind Diffusion Posterior Sampling) --- constructing multiple diffusion processes for learning the prior of each component --- which enable posterior sampling even when the operator is unknown. BlindDPS starts by initializing both the image and the operator parameter with Gaussian noise. Reverse diffusion progresses in parallel for both models, where the cross-talk between the paths are enforced from the approximate likelihood and the measurement, as can be seen in Fig.~\ref{fig:method}. With our method, both the image and the kernel starts with a coarse estimation, gradually getting closer to the ground truth as $t \rightarrow 0$ (see Fig.~\ref{fig:cover}(c)). 

In fact, our method can be thought of as a coarse-to-fine strategy naturally admitting a Gaussian scale-space representation~\cite{koenderink1984structure,lindeberg1994scale}, which can be seen as a continuous generalization of the coarse-to-fine optimization strategy that most of the optimization-based methods take~\cite{pan2016l_0,pan2017deblurring}.
Furthermore, our method is generally applicable to cases where we know the {\em structure} of the forward model a priori (e.g. convolution). To demonstrate the generality, we further show that our method can also be applied in imaging through turbulence. From our experiments, we show that the proposed method yields state-of-the-art performance while being generalizable to different inverse problems.


\section{Background}
\label{sec:background}

\paragraph{Diffusion models}
Variance preserving (VP) diffusion models (i.e. DDPM~\cite{ho2020denoising}), in the score-based persepctive~\cite{song2020score}, define the forward noising process of the data $\x(t) \triangleq \x_t,\,t \in [0, 1]$ with a linear stochastic differential equation (SDE)
\begin{align}
\label{eq:forward-sde}
    d\x = -\frac{\beta(t)}{2}\x dt + \sqrt{\beta(t)}d\w,
\end{align}
where $\beta(t)$ is the noise schedule, and $\w$ is the standard Brownian motion. One can define a proper noise schedule $\beta(t)$ such that the data distribution $\x(0) \sim p_0 = p_{\rm{data}}$ is molded into the standard Gaussian distribution $\x(1) \sim p_1 \simeq \Nc(\bm{0}, \bm{I})$.
Then, the corresponding reverse SDE is given by~\cite{anderson1982reverse}
\begin{align}
\label{eq:reverse-sde}
    d\x = \left[-\frac{\beta(t)}{2}\x - \beta(t)\nabla_{\x_t} \log p_t({\x_t})\right]dt + \sqrt{\beta(t)}d\bar\w,
\end{align}
where $\nabla_{\x_t} \log p_t({\x_t})$ is the score function, typically approximated by denoising score matching (DSM)~\cite{vincent2011connection}
\begin{equation}
\label{eq:dsm}
    \theta^* = \argmin_\theta \Ed_{t,\x_t,\x_0}\left[\|\s_\theta(\x_t, t) - \nabla_{\x_t}\log p(\x_t|\x_0)\|_2^2\right].
\end{equation}
Once trained, we can use the plug-in estimate $\nabla_{\x_t} \log p_t(\x_t) \simeq \s_\theta(\x_t,t)$ for the reverse diffusion in \eqref{eq:reverse-sde}, and solve by discretization (e.g. ancestral sampling of~\cite{ho2020denoising}), effectively sampling from the prior distribution $p(\x_0)$.

\paragraph{Diffusion posterior sampling (DPS)}
Consider the following Gaussian measurement model
\begin{align}
    p(\y|\x_0) = \Nc(\y|\Hc(\x_0), \sigma^2\bm{I}),\,\y\in\Rd^m,\,\x_0\in\Rd^n,
\end{align}
where $\y$ is the corrupted measurement, $\x_0$ is the latent image that we wish to estimate, and $\Hc$ is the forward operator. As the problem is often ill-posed, it is desirable to be able to sample from the posterior distribution $p(\x_0|\y)$. By Bayes' rule, we have for a general timestep $t$,
\begin{align}
\label{eq:bayes}
    \nabla_{\x_t} \log p(\x_t|\y) &= \nabla_{\x_t} \log p(\y|\x_t) + \nabla_{\x_t} \log p(\x_t)\\
    &\simeq \nabla_{\x_t} \log p(\y|\x_t) + \s_{\theta^*}(\x_t,t),
\label{eq:bayes_stheta}
\end{align}
where we can plug \eqref{eq:bayes_stheta} into the reverse diffusion \eqref{eq:reverse-sde} to sample from $p(\x_0|\y)$, i.e.
\begin{align}
    d\x = (-\frac{\beta(t)}{2}\x - \beta(t)[\nabla_{\x_t} \log p(\y|\x_t)\notag \\+ \s_{\theta^*}(\x_t,t)] )dt + \sqrt{\beta(t)}d\bar\w.
\label{eq:reverse-sde-posterior}
\end{align}
Note that the time-conditional log-likelihood $\log p(\y|\x_t)$ is intractable in general. However, it was shown in the work of DPS~\cite{chung2022diffusion} that we can 
use an approximation to arrive at
    \begin{align*}
    \nabla_{\x_t} \log p_t(\y|\x_t)
    &\simeq  \nabla_{\x_t} \log p(\y|\hat\x_0(\x_t)),
\end{align*}
where 
\begin{align}
    \hat\x_0(\x_t):= \frac{1}{\sqrt{{\bar\alpha(t)}}}(\x_t + (1 - {\bar\alpha(t)}) \s_{\theta^*}(\x_t,t))
\end{align}
is the denoised estimate of $\x_t$ in the VP-SDE context given by the Tweedie's formula~\cite{efron2011tweedie}.
%
Hence, one can use  the following tractable reverse SDE to sample from the posterior distribution
\begin{align}
    d\x = (-\frac{\beta(t)}{2}\x - \beta(t)[\nabla_{\x_t} \log p(\y|\hat\x_0(\x_t))\notag \\+ \s_{\theta^*}(\x_t,t)] )dt + \sqrt{\beta(t)}d\bar\w,
\label{eq:reverse-sde-DPS}
\end{align}
where we observe that $\nabla_{\x_t} \log p(\y|\hat\x_0(\x_t))$ can be efficiently computed using analytical likelihood, and backpropagation through the score function, i.e.
\begin{align*}
    \nabla_{\x_t} \log p_t(\x_t|\y) \simeq \s_{\theta^*}(\x_t) - \frac{1}{\sigma^2}\nabla_{\x_t}\|\y - \Hc(\hat\x_0(\x_t))\|_2^2.
\end{align*}
However, one should note that the method in \eqref{eq:reverse-sde-DPS} is only applicable when the forward model $\Hc$ is fixed, and hence cannot be directly used for solving {\em blind} inverse problems.

\paragraph{Blind inverse problem}
Blind inverse problems consider the case where the forward model $\Hc$ is unknown. Among them, we focus on the case where the forward operator is parameterized with $\varphib$, and we need to estimate the {\em parameter} $\varphib$. Specifically, consider the following forward model
\begin{align}
\label{eq:forward_blind_ip}
    \y = \Hc_\varphib (\x) + \n,
\end{align}
where $\varphib$ is the parameter of the forward model, $\x$ is the ground truth image, and $\n$ is some noise. Here, both $\varphib, \x$ are unknown, and should be estimated. A classical way to solve \eqref{eq:forward_blind_ip} is to optimize for the following
\begin{align}
    \min_{\x, \varphib} \quad &\frac{1}{2}\|\Hc_\varphib(\x) - \y\|^2 + R_\varphib(\varphib) + R_\x(\x),
\label{eq:opt_blind_ip}
\end{align}
where $R_\varphib(\varphib), R_\x(\x)$ are regularization functions for $\varphib, \x$, respectively, which can also be thought of as the negative log prior for each distribution, e.g. $R(\cdot) = -\log p(\cdot)$. 

\begin{figure}[t]
  \centering
    \centerline{{\includegraphics[width=1.0\linewidth]{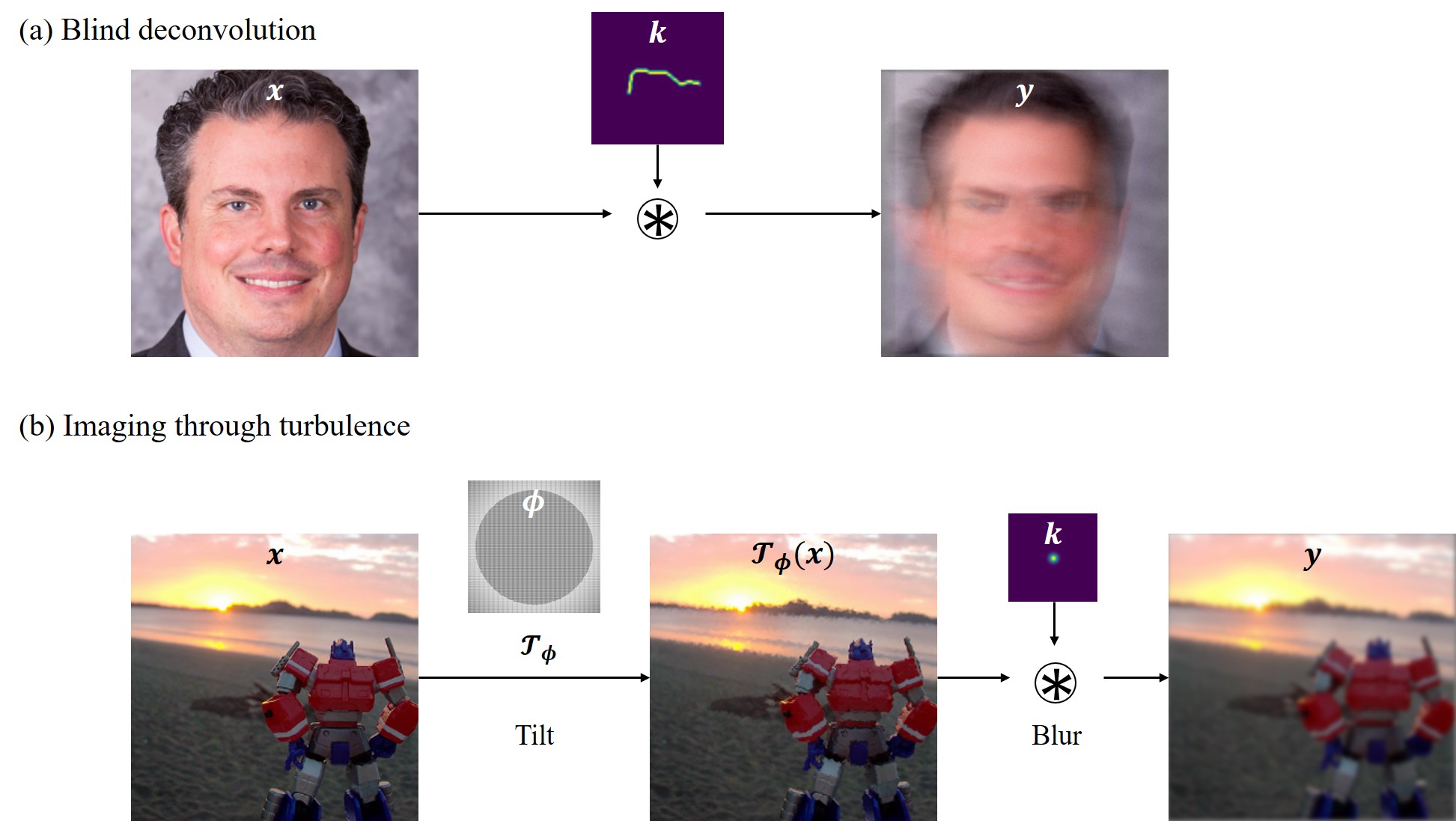}}}
  \caption{Illustration of the imaging forward model. (a) Blind deconvolution, (b) Imaging through turbulence}
  \label{fig:forward_model}
  \vspace{-0.5cm}
\end{figure}

For example, consider blind deconvolution from camera motion blur as illustrated in Fig.~\ref{fig:forward_model}(a). The forward model reads 
\begin{align}
\label{eq:motion}
\y = \kb \ast \x + \n,
\end{align}
where
$\kb$ is the blur kernel, corresponding to the parameter $\varphib$. 
On the other hand, although the ``real'' forward model for atmospheric turbulence is rarely directly used in practice due to the highly complicated nature of the wave propagation theory, the tilt-blur model is often used~\cite{chan2022tilt,shimizu2008super,chak2018subsampled}, as the model is simple but fairly accurate. Specifically, the visualization of such imaging process is shown in Fig.~\ref{fig:forward_model}(b), which can be mathematically
described by
\begin{align}
\label{eq:turblence}
\yb = \kb \ast \Tc_{\bm{\phi}}(\x) +\n,
\end{align}
where $\Tc$ is the tilt operator parameterized by the tilt vector field $\bm{\phi}$. 
To remove the scale ambiguity between the kernel and image, the magnitude and the polarity
constraints of kernels 
are often used: 
\begin{align}
     \boldsymbol{1}^T \kb = 1 , \kb \succeq 0 .
\label{eq:const_opt_bd}
\end{align}
Then, the success of the optimization algorithm \eqref{eq:opt_blind_ip} with the forward models
\eqref{eq:motion} or \eqref{eq:turblence} under the constraint  \eqref{eq:const_opt_bd}
 depends on two factors: 1) How closely the prior-imposing functions $R_{\{\x, \kb\}}$ estimate the true prior, and 2) how well the optimization procedure finds the minimum value. Conventional methods are sub-optimal in both aspects. First, the prior (e.g. sparsity~\cite{pan2016l_0}, dark channel~\cite{pan2017deblurring}, implicit from deep networks~\cite{ren2020neural}) functions do not fully represent the {\em true} prior. Second, the optimization process is unstable and hard to tune. For instance, \cite{pan2016l_0,pan2017deblurring} requires different weighting parameters {\em per image}, and often fails during the abrupt changes in the stage transition during coarse-to-fine optimization strategy. In section~\ref{sec:blindDPS}, we show that our method can solve both of these problems.

%

\section{BlindDPS}
\label{sec:blindDPS}

In DPS~\cite{chung2022diffusion}, the authors used the diffusion prior for $R_\x$ by training a score function that models $\nabla_\x \log p(\x)$. As for blind inverse problems, a prior model for the parameter $p(\varphib)$ should also be specified. In this regard, our proposal is to use the diffusion prior also for the forward model parameter by estimating $\nabla_\varphib \log p(\varphib)$. With such choice, one can model a much more accurate prior for the parameters compared to the conventional choices. In the following, we detail on how to build our method {\em BlindDPS}, focusing on blind deconvolution. The method for imaging through turbulence can be derived in a completely analogous fashion, where the details can be found in Supplementary section~\ref{sec:supp_blinddps_itt}.

\noindent
\textbf{Key idea.~}
In blind deblurring (deconvolution), the probabilistic forward model is specified as follows
\begin{align}
    p(\y|\x_0, \kb_0) := \Nc(\y| \kb_0 \ast \xb_0, \sigma^2 \Ib),
\end{align}
where $\kb_0$ is the random variable of the convolution kernel. As $\x_0$ and $\kb_0$ are independent, the posterior probability is given as
\begin{align}
\label{eq:posterior_blind_deconv}
    p(\x_0, \kb_0|\y) \propto p(\y|\x_0, \kb_0)p(\x_0)p(\kb_0).
\end{align}
Note that our aim is to use implicit diffusion priors for both $p(\x_0)$ and $p(\kb_0)$ through their score functions. One can easily take pre-trained score functions for the image. Similarly, the score function for the kernel can also be estimated from standard DSM~\eqref{eq:dsm} to get $\s_{\theta^*}^k(\kb, t) \simeq \nabla_{\kb_t} \log p_t(\kb_t)$. Note that performing DSM to achieve $\s_{\theta^*}^k$ costs much less than training the image score function $\s_{\theta^*}^i$, as the distribution is much simpler, and the dimensionality of the vector $\kb$ is also sufficiently smaller than $\x$.

On the other hand, again from the independence of $\x_0$ and $\kb_0$, we are able to construct two separate reverse diffusion processes of identical form:
\begin{align*}
    d\x &= \left[-\frac{\beta(t)}{2}\x - \beta(t)\nabla_{\x_t} \log p(\x_t)\right]dt + \sqrt{\beta(t)}d\bar\w,\\
    d\kb &= \left[-\frac{\beta(t)}{2}\kb - \beta(t)\nabla_{\kb_t} \log p(\kb_t)\right]dt + \sqrt{\beta(t)}d\bar\w.
\end{align*}
Note that the two reverse SDEs are only able to sample from the marginals --- $p(\x_0)$, $p(\kb_0)$. However, one can define the dependency between $\x, \y,$ and $\kb$ from the posterior probability.
Using Bayes' rule in \eqref{eq:posterior_blind_deconv} for general $t$, we have
\begin{align*}
    \nabla_{\x_t} \log p(\x_t, \kb_t|\y) &= \nabla_{\x_t} \log p(\y|\x_t, \kb_t) + \nabla_{\x_t} \log p(\x_t),\\
    \nabla_{\kb_t} \log p(\x_t, \kb_t|\y) &= \nabla_{\kb_t} \log p(\y|\x_t, \kb_t) + \nabla_{\kb_t} \log p(\kb_t).
\end{align*}
Here, in order to estimate the time-conditional log-likelihood $\log p(\y|\x_t, \kb_t)$ which is intractable in general, we need the following result:
\begin{restatable}[]{theorem}{ddlaplace}
    \label{thm:ddlaplace}
Under the same conditions in~\cite{chung2022diffusion}, we have
    \begin{align*}
    \nabla_{\x_t} \log p_t(\y|\x_t, \kb_t)
    &\simeq  \nabla_{\x_t} \log p(\y|\hat\x_0(\x_t), \hat\kb_0(\kb_t)) \\
    \nabla_{\kb_t} \log p_t(\y|\x_t, \kb_t)
    &\simeq  \nabla_{\kb_t} \log p(\y|\hat\x_0(\x_t), \hat\kb_0(\kb_t)).
\end{align*}
\end{restatable}
\begin{remark}
Our theorem holds as long as $\x_t$, $\kb_t$ are independent. Note that the theorem can be further generalized to handle more random variables whenever the independence between the variables is established. In other words, we can construct arbitrary many diffusion procedures for each component of the forward model, which can be solved analogous to the approximation proposed in Theorem~\ref{thm:ddlaplace}. This result will be useful when we solve the problem of imaging through turbulence in in Supplementary section~\ref{sec:supp_blinddps_itt}.
\label{rmk:generalize}
\end{remark}
\noindent
Using Theorem~\ref{thm:ddlaplace}, we finally arrive at the following reverse SDEs
\begin{align}
\label{eq:reverse-sde-x}
    d\x &= (-\frac{\beta(t)}{2}\x - \beta(t)[\nabla_{\x_t} \log p(\y|\hat\x_0(\x_t), \hat\kb_0(\kb_t))\notag \\&+ \s_{\theta^*}^i(\x_t,t)] )dt + \sqrt{\beta(t)}d\bar\w,\\
    d\kb &= (-\frac{\beta(t)}{2}\kb - \beta(t)[\nabla_{\kb_t} \log p(\y|\hat\x_0(\x_t), \hat\kb_0(\kb_t))\notag \\&+ \s_{\theta^*}^k(\kb_t,t)] )dt + \sqrt{\beta(t)}d\bar\w.
\label{eq:reverse-sde-k}
\end{align}
The system of equations \eqref{eq:reverse-sde-x},\eqref{eq:reverse-sde-k} are now numerically solvable as the gradient of the log likelihood is analytically tractable. Specifically, for the Gaussian measurement, we have
\begin{align}
    \nabla_{\x_t} \log p(\y|\hat\x_0, \hat\kb_0) = - \frac{1}{\sigma^2} \nabla_{\x_t} \|\y - \hat\kb_0 \ast \hat\x_0\|_2^2.
\end{align}
Combined with the ancestral sampling steps~\cite{ho2020denoising}, our algorithm for posterior sampling of blind deblurring is formally given in Algorithm~\ref{alg:db}. Here, note that we choose to take static step size times the gradient of the norm instead of taking time-dependent step sizes times the gradient of the squared norm, as it was shown to be effective despite its simplicity~\cite{chung2022diffusion}. Furthermore, in order to impose the usual condition \eqref{eq:const_opt_bd}, we define a set $C := \{\kb|\boldsymbol{1}^T \kb = 1 , \kb \succeq 0\}$, and project onto the set through $\Pc_C(\hat\kb_0)$ in  Algorithm~\ref{alg:db}, after the estimation of $\hat\kb_0$ at each intermediate step. For visual illustration of the proposed method, see Fig.~\ref{fig:method}.

\begin{algorithm}[!t]
\caption{BlindDPS --- Blind Deblurring}
\begin{algorithmic}[1]
\Require $N$, $\y$, $\alpha,  {\{\tilde\sigma_i\}_{i=1}^N}, \lambda, {R_\kb}(\cdot)$
\State $\x_N,\kb_N \sim \Nc(\bm{0}, \bm{I})$
\For{$i=N-1$ {\bfseries to} $0$}
 \State{{$\hat\s^i \gets \s_{\theta^*}^i(\x_i, i)$}}
 \State{{$\hat\s^k \gets \s_{\theta^*}^k(\kb_i, i)$}}
 \State{{$\hat\x_0 \gets \frac{1}{\sqrt{\bar\alpha_i}}(\x_i + \sqrt{1 - \bar\alpha_i}\hat\s^i)$}}
 \State{{$\hat\kb_0 \gets \frac{1}{\sqrt{\bar\alpha_i}}(\kb_i + \sqrt{1 - \bar\alpha_i}\hat\s^k)$}}
 \State{{$\hat\kb_0 \gets \Pc_C(\hat\kb_0)$}}
 \State{$\z_i,\z_k \sim \Nc(\bm{0}, \bm{I})$}
 \State{$\x'_{i-1} \gets \frac{\sqrt{\alpha_i}(1-\bar\alpha_{i-1})}{1 - \bar\alpha_i}\x_i + \frac{\sqrt{\bar\alpha_{i-1}}\beta_i}{1 - \bar\alpha_i}\hat\x_0 +  {\tilde\sigma_i \z_i}$}
 \State{$\kb'_{i-1} \gets \frac{\sqrt{\alpha_i}(1-\bar\alpha_{i-1})}{1 - \bar\alpha_i}\kb_i + \frac{\sqrt{\bar\alpha_{i-1}}\beta_i}{1 - \bar\alpha_i}\hat\kb_0 +  {\tilde\sigma_i \z_k}$}
 \State{$\x_{i-1} \gets \x'_{i-1} - \alpha\nabla_{\x_i}\|\y - \hat\kb_0 \ast \hat\xb_0\|_2$}
 \State{$\Lc_\kb \gets \|\y - \hat\kb_0 \ast \hat\xb_0\|_2 + {\lambda R_\kb(\hat\kb_0)}$}
 \State{$\kb_{i-1} \gets \kb'_{i-1} - \alpha\nabla_{\kb_i}\Lc_\kb$}
\EndFor
\State {\bfseries return} $\x_0, \kb_0$
\end{algorithmic}\label{alg:db}
\end{algorithm}

\noindent
\textbf{Augmenting diffusion prior with sparsity.~}
Implementing \eqref{eq:reverse-sde-x},\eqref{eq:reverse-sde-k} directly induces fairly stable results with the correct choice of $\alpha$. Here, we go a step further and adopt a lesson from the classic literature. As we often wish to estimate blur kernels that are sparse, we promote sparsity {\em only} to the kernel that we are estimating by augmenting the diffusion prior with $\ell_0$/$\ell_1$ regularization. The minimization strategy for the kernel then becomes
\begin{align}
    \kb_{i-1} = \kb'_{i-1} - \alpha\left(\|\y - \hat\kb_0 \ast \hat\x_0\|_2 + \lambda {R_\kb}(\hat\kb_0)\right),
\label{eq:sparsity-reg}
\end{align}
where $\lambda$ is the regularization strength, and the choice of {$R_\kb(\cdot) := \ell_0$/$\ell_1$} regularization depends on the type of the dataset. With such augmentation, reconstruction can be further stabilized.

\noindent
\textbf{Interpretation in Gaussian scale-space.~}
(Gaussian) Scale-space theory~\cite{lindeberg1994scale} states that one can represent signals in multiple scales by gradually convolving with Gaussian filters. As adding Gaussian noise to random vectors in the forward pass of the diffusion has a dual relation in the density domain (i.e. convolution with Gaussian kernels), one can think of the diffusion process as a realization of one such process. Thus, the reverse diffusion process can be interpreted as a  coarse-to-fine synthesis evolving through the Gaussian scale-space, which is most visible by visualizing $\hat\x_0(\x_t), \hat\kb_0(\kb_t)$ when evolving through $t = 1 \rightarrow 0$ (see Fig.~\ref{fig:cover}(c)).

For blind deconvolution problems, in order to achieve optimal quality, it is a standard practice to start the optimization process at a coarse scale by down-sampling, and sequentially upsample with a pre-determined schedule to refine the estimates~\cite{pan2016blind,pan2016l_0}. However, the discretized schedule is typically abrupt (e.g. \cite{pan2016blind,pan2016l_0} uses 8 discretization) and ad-hoc. On the other hand, by using the reverse diffusion process, we are granted with a natural, smooth schedule of evolution, which can be thought of as a continuous generalization of the coarse-to-fine reconstruction strategy. This could be another reason why the proposed method is able to dramatically outperform the conventional methods.

\begin{figure*}[t]
  \centering
    \centerline{{\includegraphics[width=1.0\linewidth]{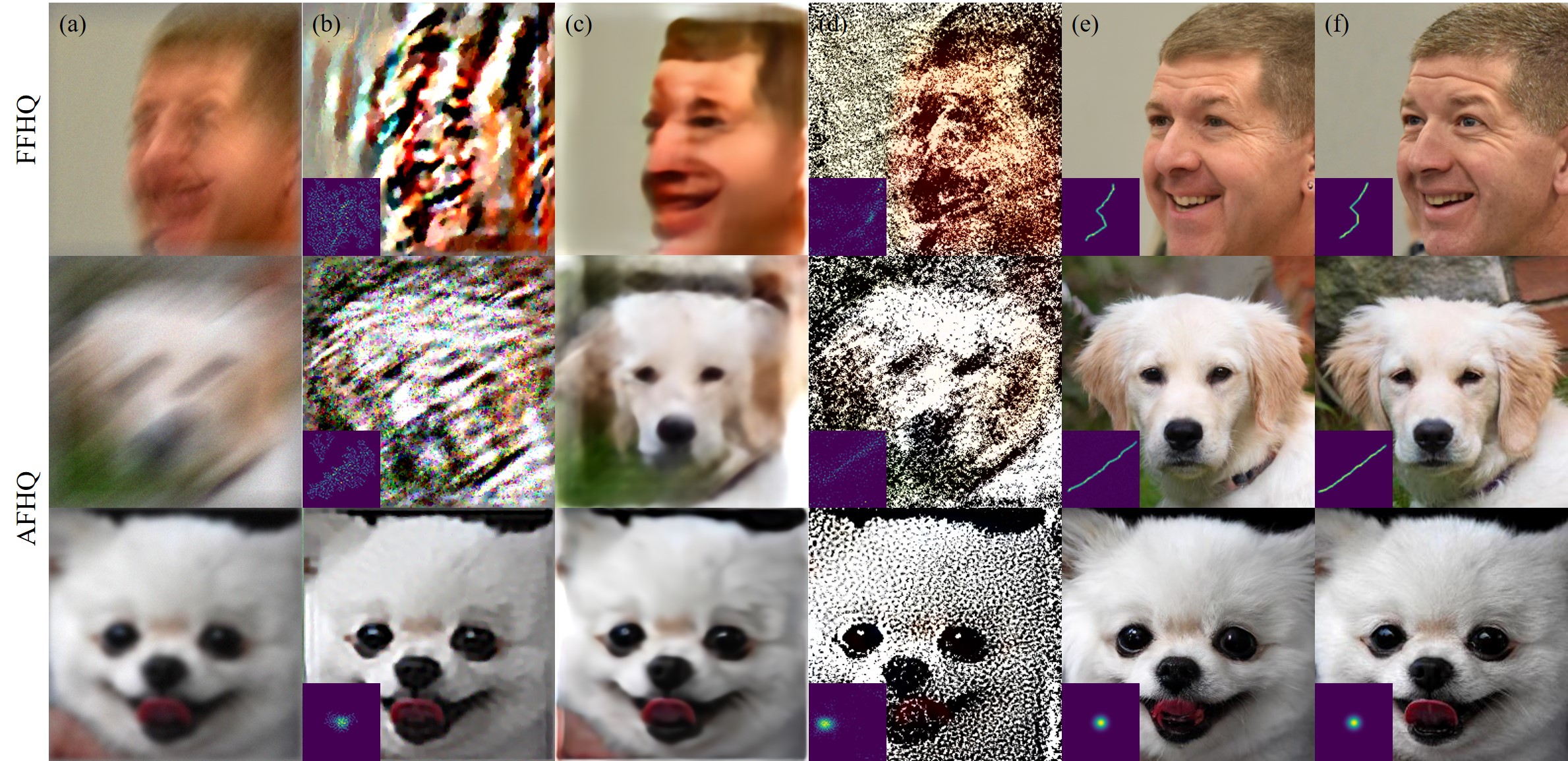}}}
  \caption{{Blind deblurring results. (row 1): FFHQ 256$\times$256 motion deblurring, (row 2): AFHQ 256$\times$256 motion deblurring. (row 3): AFHQ 256$\times$256 Gaussian deblurring. (a) Measurement, (b) Pan-DCP~\cite{pan2017deblurring}, (c) MPRNet~\cite{zamir2021multi}, (d) SelfDeblur~\cite{ren2020neural}, (e) BlindDPS (\textbf{ours}), (f) Ground truth. For (c), kernel not shown as the method only estimate images.}}
  \label{fig:results_motion_deblur}
\end{figure*}

\begin{table*}[]
\centering
\setlength{\tabcolsep}{0.2em}
\resizebox{0.8\textwidth}{!}{
\begin{tabular}{lllllll@{\hskip 15pt}llllll}
\toprule
{} & \multicolumn{6}{c}{\textbf{FFHQ} ($\bf 256 \times 256$)} & \multicolumn{6}{c}{\textbf{AFHQ} ($\bf 256 \times 256$)} \\
\cmidrule(lr){2-7}
\cmidrule(lr){8-13}
{} & \multicolumn{3}{c}{\textbf{Motion}} & \multicolumn{3}{c}{\textbf{Gaussian}} &
\multicolumn{3}{c}{\textbf{Motion}} & \multicolumn{3}{c}{\textbf{Gaussian}} \\
\cmidrule(lr){2-4}
\cmidrule(lr){5-7}
\cmidrule(lr){8-10}
\cmidrule(lr){11-13}
{\textbf{Method}} & {FID $\downarrow$} & {LPIPS $\downarrow$} & {PSNR $\uparrow$} & {FID $\downarrow$} & {LPIPS $\downarrow$} & {PSNR $\uparrow$} & {FID $\downarrow$} & {LPIPS $\downarrow$} & {PSNR $\uparrow$} & {FID $\downarrow$} & {LPIPS $\downarrow$} & {PSNR $\uparrow$}\\
\midrule
BlindDPS~\textcolor{trolleygrey}{(ours)} & \textbf{29.49} & \textbf{0.281} & \textbf{22.24} & \textbf{27.36} & \textbf{0.233} & \textbf{24.77} & \textbf{23.89} & \textbf{0.338} & \textbf{20.92} & \textbf{20.54} & \textbf{0.287} & \textbf{23.63}\\
\cmidrule(l){1-13}
SelfDeblur~\cite{ren2020neural} & 270.0 & 0.717 & 10.83 & 235.4 & 0.686 & 11.36 & 300.5 & 0.768 & 9.081 & 172.2 & 0.662 & 11.53\\
MPRNet~\cite{zamir2021multi} & \underline{111.6} & \underline{0.434} & 17.40 & 95.12 & \underline{0.337} & \underline{20.75} & \underline{131.8} & \underline{0.521} & 14.85 & \underline{46.19} & \underline{0.366} & 20.51\\
DeblurGANv2~\cite{kupyn2019deblurgan} & 220.7 & 0.571 & \underline{17.75} & 185.5 & 0.529 & 19.69 & 186.2 & 0.597 & \underline{17.64} & 86.87 & 0.523 & 20.29\\
Pan-DCP~\cite{pan2017deblurring} & 214.9 & 0.520 & 15.41 & \underline{92.70} & 0.393 & 20.50 & 214.0 & 0.704 & 11.87 & 57.14 & 0.392 & \underline{20.97}\\
Pan-$\ell_0$~\cite{pan2016l_0} & 242.6 & 0.542 & 15.53 & 109.1 & 0.415 & 19.94 & 235.0 & 0.627 & 15.34 & 62.76 & 0.395 & 21.41\\
\bottomrule
\end{tabular}
}
\caption{
Quantitative evaluation (FID, LPIPS, PSNR) of blind deblurring task on FFHQ and AFHQ. \textbf{Bold}: Best, \underline{under}: second best.
}
\vspace{-0.5cm}
\label{tab:deblur_quantitative}
\end{table*}

\begin{table}[]
\centering
\setlength{\tabcolsep}{0.2em}
\resizebox{0.45\textwidth}{!}{
\begin{tabular}{lllll@{\hskip 15pt}llll}
\toprule
{} & \multicolumn{4}{c}{\textbf{FFHQ} ($\bf 256 \times 256$)} & \multicolumn{4}{c}{\textbf{AFHQ} ($\bf 256 \times 256$)} \\
\cmidrule(lr){2-5}
\cmidrule(lr){6-9}
{} & \multicolumn{2}{c}{\textbf{Motion}} & \multicolumn{2}{c}{\textbf{Gaussian}} &
\multicolumn{2}{c}{\textbf{Motion}} & \multicolumn{2}{c}{\textbf{Gaussian}} \\
\cmidrule(lr){2-3}
\cmidrule(lr){4-5}
\cmidrule(lr){6-7}
\cmidrule(lr){8-9}
{\textbf{Method}} & {MSE $\downarrow$} & {MNC $\uparrow$} & {MSE $\downarrow$} & {MNC $\uparrow$} & {MSE $\downarrow$} & {MNC $\uparrow$} & {MSE $\downarrow$} & {MNC $\uparrow$} \\
\midrule
BlindDPS~\textcolor{trolleygrey}{(ours)} & \textbf{0.003} & \textbf{0.955} & \textbf{0.000} & \textbf{0.995} & \textbf{0.003} & \textbf{0.930} & \textbf{0.001} & \textbf{0.991} \\
\cmidrule(l){1-9}
SelfDeblur~\cite{ren2020neural} & 0.021 & 0.323 & 0.020 & 0.266 & 0.021 & 0.268 & 0.020 & 0.272\\
Pan-DCP~\cite{pan2017deblurring} & \underline{0.020} & 0.425 & \underline{0.016} & 0.478 & \underline{0.020} & 0.365 & 0.016 & 0.481\\
Pan-$\ell_0$~\cite{pan2016l_0} & \underline{0.020} & \underline{0.454} & \underline{0.016} & \underline{0.518} & \underline{0.020} & \underline{0.398} & \underline{0.015} & \underline{0.517}\\
\bottomrule
\end{tabular}
}
\caption{
Quantitative evaluation (MSE, MNC~\cite{hu2012good}) of kernel estimation on FFHQ and AFHQ. \textbf{Bold}: Best, \underline{under}: second best.
}
\vspace{-0.5em}
\label{tab:kernel_quantitative}
\end{table}
\section{Experiments}
\label{sec:exp}

\subsection{Experimental setup}

\noindent
\textbf{Blind deblurring.~}
For blind deblurring, we conduct experiments on FFHQ 256$\times$256~\cite{karras2019style}, and AFHQ-dog 256$\times$256~\cite{choi2020stargan} dataset on \{motion, Gaussian\}-deblurring. We choose 1k validation set for FFHQ, and use 500 test sets for AFHQ-dog. We leverage pre-trained score functions, as in the experimental setting of~\cite{chung2022come}. We train the score function on 60k generated blur kernels of size $64\times 64$ (both Gaussian and motion\footnote{\url{https://github.com/topics/motion-blur}}) for 3M steps with a small U-Net~\cite{dhariwal2021diffusion}. For testing, motion blur kernel is randomly generated with with intensity 0.5 following~\cite{chung2022diffusion}, and the standard deviation of the gaussian kernels is set to 3.0.  {Step size for Algorithm~\ref{alg:db} is set to $\alpha = 0.3$ for both FFHQ/AFHQ. We choose $R_\kb(\cdot) = \ell_1, \lambda=1.0$ for FFHQ, and $R_\kb(\cdot) = \ell_0, \lambda = 5.0$ for AFHQ.}

\noindent
\textbf{Imaging through turbulence.~}
For imaging through turbulence, we conduct experiments with FFHQ 256$\times$256, and ImageNet 256$\times$256~\cite{deng2009imagenet}, with pre-trained ImageNet score function taken from~\cite{dhariwal2021diffusion}. The score function for kernel blur is taken from the blind deblurring experiment, and the score function for the tilt map is trained with 50k randomly generated tilt maps following~\cite{chak2018subsampled}. The point spread function (PSF) is assumed to be a Gaussian with standard deviation of 4.0, 2.0 for FFHQ, ImageNet, respectively (size 64$\times$64). For both blind inverse problems, we add Gaussian measurement noise with $\sigma = 0.02$. {Step size is set to $\alpha = 0.3$.} Full details on experimental setup can be found in supplementary section~\ref{sec:exp_details}. 

\noindent
\textbf{Evaluation.~}
We use three metrics---Frechet inception distance (FID), learned Perceptual Image Patch Similarity (LPIPS), and peak signal-to-noise-ratio (PSNR)---for quantitatively measuring the performance of the image reconstruction. For kernel estimation, we use mean-squared-error (MSE), and maximum of normalized convolution (MNC)~\cite{hu2012good}, which is computed by
\begin{align}
\label{eq:mnc}
    {\rm MNC} := \max\left(\frac{\tilde\kb \ast \kb^*}{\|\tilde\kb\|_2 \|\kb^*\|_2}\right),
\end{align}
where $\tilde\kb, \kb^*$ are the estimated, and the ground truth kernels, respectively.

\noindent
\textbf{Comparison methods.~}
For blind deblurring, we compare the reconstruction performance of BlindDPS against state-of-the-art methods. Specifically, we choose MPRNet~\cite{zamir2021multi} and DeblurGANv2~\cite{kupyn2019deblurgan} as supervised learning-based baselines that are incapable of kernel estimation, but work through amortized inference. We also compare our method against SelfDeblur~\cite{ren2020neural}, which leverages deep image prior (DIP) for estimating both the kernel and the image. For optimization-based methods, we use Pan-dark channel prior (Pan-DCP)~\cite{pan2017deblurring}, and Pan-$\ell_0$~\cite{pan2016l_0}. For imaging through turbulence, we use MPRNet~\cite{zamir2021multi}, DeblurGANv2~\cite{kupyn2019deblurgan}, and TSR-WGAN~\cite{jin2021neutralizing} as comparison methods that are based on supervised traning. We also compare against ILVR~\cite{choi2021ilvr}, which is a diffusion model-based method that is capable of restoring images from low resolution.

\subsection{Results}

\noindent
\textbf{Blind deblurring.~}
Motion deblurring results are presented in Fig.~\ref{fig:cover}(a) and Fig.~\ref{fig:results_motion_deblur}. As our setting for motion deblurring imposes a rather aggressive degradation with a large blur kernel, most of the prior arts fail catastrophically, not being able to generate a feasible solution. In contrast, our method accurately captures both the kernel and the image with sharpness. Similar trend can be seen for Gaussian deblurring presented in the third row of
 Fig.~\ref{fig:results_motion_deblur}. Other methods fall far short of BlindDPS in the sense that they either produce reconstructions that are blurry with inaccurate blur kernel estimation, or fails dramatically (e.g. SelfDeblur). Furthermore, the proposed method establishes the state-of-the-art in all quantitative metrics, which can be seen in Table~\ref{tab:deblur_quantitative} and Table~\ref{tab:kernel_quantitative}.

\begin{figure*}[t]
  \centering
    \centerline{{\includegraphics[width=1.0\linewidth]{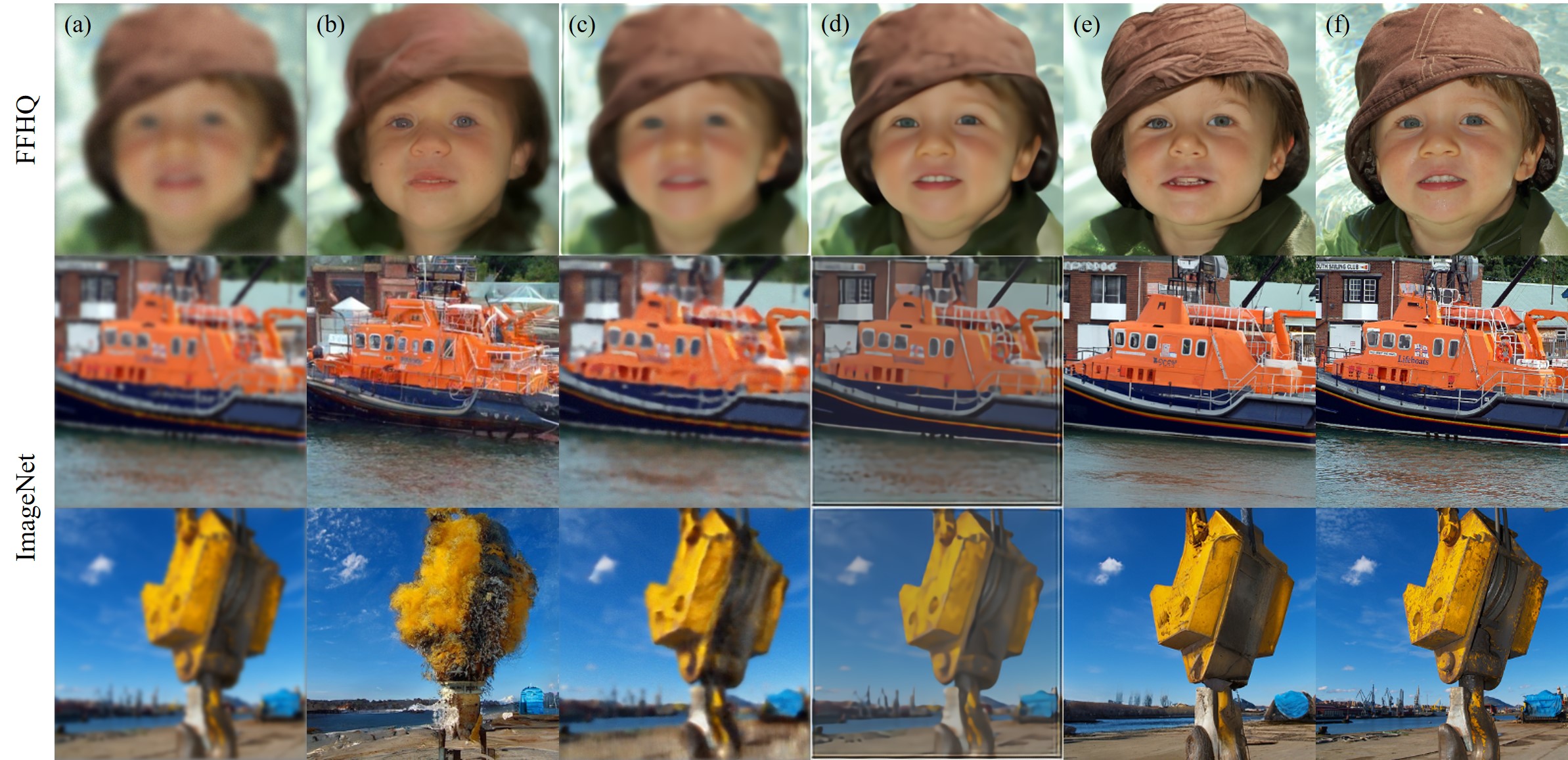}}}
  \caption{{Reconstruction of imaging through turbulence. (row 1): FFHQ 256$\times$256, (row 2-3): ImageNet 256$\times$256. (a) Measurement, (b) ILVR~\cite{choi2021ilvr}, (c) MPRNet~\cite{zamir2021multi}, (d) TSR-WGAN~\cite{jin2021neutralizing}, (e) BlindDPS (\textbf{ours}), (f) Ground truth.}}
  \label{fig:results_turbulence}
  \vspace{-0.5cm}
\end{figure*}

\noindent
\textbf{Imaging through turbulence.~}
\begin{table}[t]
\centering
\setlength{\tabcolsep}{0.2em}
\resizebox{0.45\textwidth}{!}{
\begin{tabular}{llll@{\hskip 15pt}llll}
\toprule
{} & \multicolumn{3}{c}{\textbf{FFHQ} ($\bf 256 \times 256$)} & \multicolumn{3}{c}{\textbf{ImageNet} ($\bf 256 \times 256$)} \\
\cmidrule(lr){2-4}
\cmidrule(lr){5-7}
{\textbf{Method}} & {FID $\downarrow$} & {LPIPS $\downarrow$} & {PSNR $\uparrow$} & {FID $\downarrow$} & {LPIPS $\downarrow$} & {PSNR $\uparrow$}\\
\midrule
BlindDPS~\textcolor{trolleygrey}{(ours)} & \textbf{27.35} & \textbf{0.247} & \underline{24.49} & \textbf{51.25} & \textbf{0.341} & 19.59\\
\cmidrule(l){1-7}
TSR-WGAN~\cite{jin2021neutralizing} & \underline{58.30} & \underline{0.258} & \textbf{26.29} & 69.80 & \underline{0.369} & 17.67\\
ILVR~\cite{choi2021ilvr} & 65.50 & 0.370 & 21.48 & 85.21 & 0.494 & 18.09\\
MPRNet~\cite{zamir2021multi} & 116.2 & 0.411 & 19.68 & 78.24 & 0.421 & \underline{20.34}\\
DeblurGANv2~\cite{kupyn2019deblurgan} & 225.9 & 0.561 & 18.40 & \underline{60.31} & 0.393 & \textbf{21.56}\\
\bottomrule
\end{tabular}
}
\caption{
Quantitative evaluation (FID, LPIPS, PSNR) of imaging through turbulence task on FFHQ and ImageNet. \textbf{Bold}: Best, \underline{under}: second best.
}
\label{tab:turbulence_quantitative}
\vspace{-0.2em}
\end{table}
We show the reconstruction results in Fig.~\ref{fig:cover}(b) and Fig.~\ref{fig:results_turbulence}, with quantitative metrics in Table~\ref{tab:turbulence_quantitative}. Consistent with the results from blind deblurring, BlindDPS outperforms the comparison methods in most cases, effectively removing both the blur and the tilt from the measurement. Notably, our method outperforms {\em all} other methods by large margins on perceptual metrics (i.e. FID, LPIPS). For PSNR, the proposed method often slightly underperforms against supervised learning approaches, which is to be expected, as for reconstructions from heavy degradations, retrieving the high-frequency details often penalizes such distortion metrics~\cite{blau2018perception}.

\subsection{Ablation studies}
We perform two ablation studies to verify our design choices: 1) using the diffusion prior for the forward model parameters, and 2) augmenting the diffusion prior with the sparsity prior. Details on the experimental setup along with further analysis can be found in Supplementary section~\ref{sec:detailed_ablation_studies}.

\begin{figure}[t]
  \centering
  \centerline{{\includegraphics[width=1.0\linewidth]{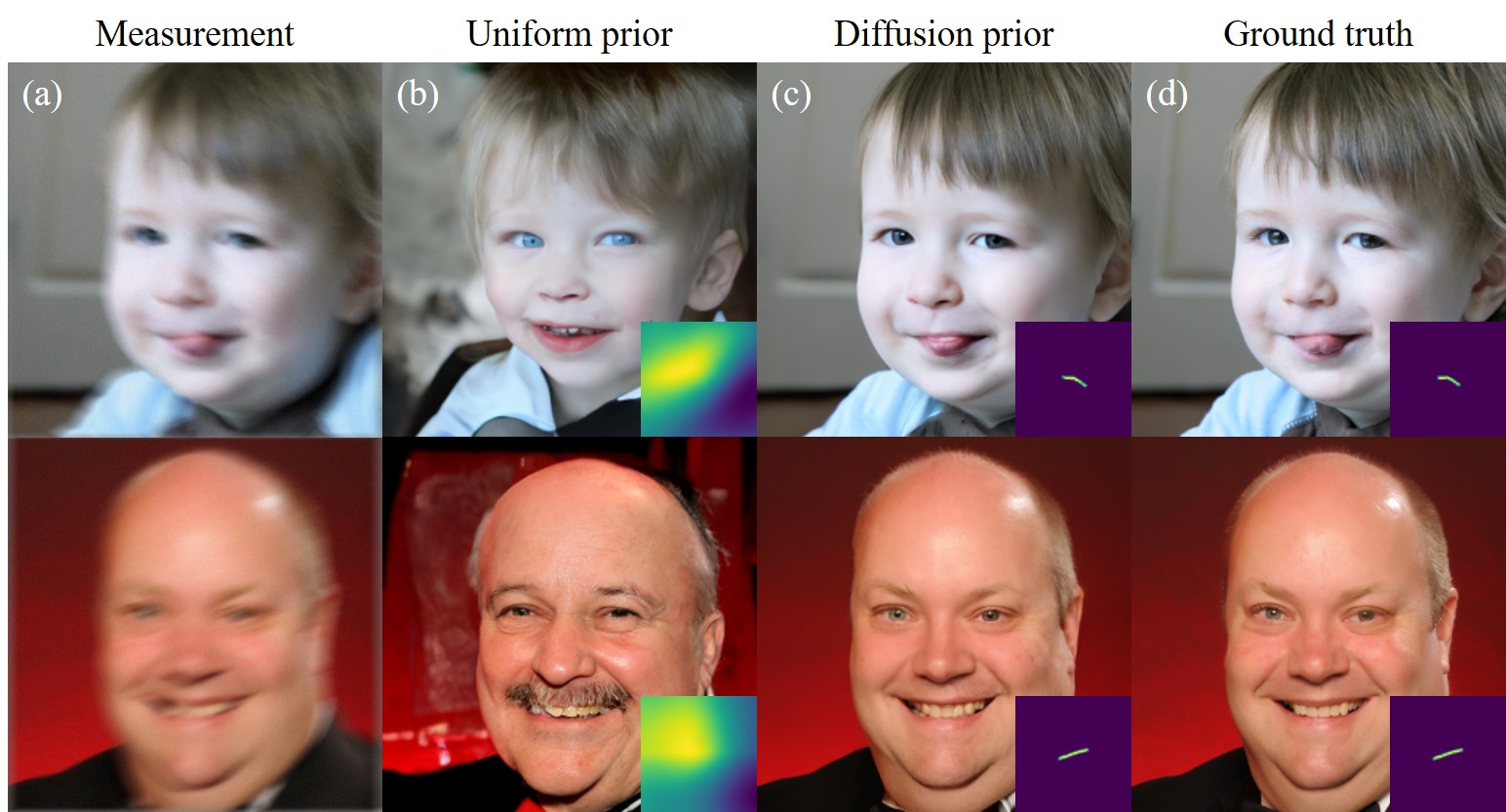}}}
  \caption{Ablation study: uniform prior vs. diffusion prior. (a) Measurement, (b) uniform prior, (c) diffusion prior, (d) ground truth.}
  \vspace{-0.5cm}
  \label{fig:ablation_uniformprior}
\end{figure}

\noindent
\textbf{Diffusion prior for the forward model.~}
One may question why the score function for the kernel is necessary in the first place, since one could also estimate the kernel solely through gradient descent using the gradient of the likelihood. In fact, this corresponds to using the uniform prior for the kernel distribution, which we compare against the proposed diffusion prior (BlindDPS) in Fig.~\ref{fig:ablation_uniformprior}. We clearly see that using the uniform prior yields heavily distorted result, with poorly estimated kernel. From this experiment, we observe that using another diffusion process specifically for the forward model is crucial for the performance.

\noindent
\textbf{Effect of sparsity regularization.~}
\begin{table}[t]
\centering
\setlength{\tabcolsep}{0.2em}
\resizebox{0.40\textwidth}{!}{
\begin{tabular}{lllll@{\hskip 15pt}llll}
\toprule
{} & \multicolumn{4}{c}{\textbf{Motion}} & \multicolumn{4}{c}{\textbf{Gaussian}} \\
\cmidrule(lr){2-5}
\cmidrule(lr){6-9}
{$\lambda$} & {0.0} & {0.1} & {1.0} & {5.0} & {0.0} & {0.1} & {1.0} & {5.0}\\
\midrule
MNC $\uparrow$ & 0.929 & 0.956 & 0.958 & 0.959 & 0.996 & 0.997 & 0.996 & 0.997 \\
MSE $\downarrow$ & 0.004 & 0.002 & 0.002 & 0.002 & 0.000 & 0.000 & 0.000 & 0.000 \\
\bottomrule
\end{tabular}
}
\caption{
Ablation study: effect of sparsity regularization in blind deconvolution.
}
\vspace{-0.5cm}
\label{tab:ablation_sparsity_kernel}
\end{table}
One design choice made in BlindDPS is the additional sparsity regularization applied to kernels. Here, we analyze the effect of such regularization. In Table~\ref{tab:ablation_sparsity_kernel}, we report on quantitative metrics for the kernel, depending on the regularization weight $\lambda$. Clearly, setting $\lambda = 0.0$ induces inferior performance especially for motion deblurring. When setting $\lambda \geq 0.1$ however, we can see that one can achieve good performance regardless of the chosen weight value. As diffusion priors have been shown to have surprisingly high generalization capacity~\cite{jalal2021robust, chung2022score}, we choose a mild weight value of $\lambda = 1.0$, which gives visually appealing results without down-weighting the influence of diffusion priors too much.

\section{Discussion and Related Works}
\label{sec:discussion_related_works}

This work follows the line of endeavors to develop methods that can solve inverse problems through diffusion models. Methods that are based on iterative projection onto convex sets (POCS) were the first to be developed, iterating between the denoising step, and the projection step~\cite{song2020score,chung2022come,choi2021ilvr,song2021solving,chung2022score}. Methods that attempt to approximate posterior sampling via annealed Langevin dynamics (ALD)~\cite{jalal2021robust}, and singular value decomposition (SVD)~\cite{kawar2022denoising} were proposed, with the latter showing particular robustness to noisy measurements.

The trend recently shifted towards leveraging the denoised estimate via Tweedie's formula at the intermediate steps under various names --- manifold constrained gradient (MCG)~\cite{chung2022improving}, gradient guidance~\cite{ho2022video}, and reconstruction-based method~\cite{kawar2022jpeg}. Diffusion posterior sampling (DPS)~\cite{chung2022diffusion} is the method that is the most similar to ours, showing that such method is an approximation of the posterior sampling process. However, none of the methods so far considered blind inverse problem, and to the best of our knowledge, we are the first to show that posterior sampling with diffusion scales to blind settings.

\noindent
\textbf{Limitations and future directions.~} As BlindDPS performs joint minimization on multiple factors (e.g. kernel, tilt-map, image), it is typically less robust than the non-blind reconstruction scheme. At times, the solution diverges when the parameters are incorrectly tuned. For imaging through turblence, it is often the case where the tilt map is incorrectly estimated whereas the kernel and the ground truth image are accurately estimated.
Furthermore, as we train and use specified diffusion score functions for each of the component, inference speed is delayed, due to the additional forward/backward passes through the newly involved score functions. When the forward functional involves estimating additional parameters, the number of score functions required will scale linearly, not being efficient with complex functional forms. Finally, we note that our method is yet to solve the {\em truly blind} case, where we do not know the functional form of the forward mapping. Solving the truly blind case would be an interesting direction of future studies.

\section{Conclusion}
\label{sec:conclusion}

In this work, we proposed BlindDPS, a framework for solving blind inverse problems by jointly estimating the parameters of the forward measurement operator and the image to be reconstructed. We theoretically show how we can construct a system of reverse SDEs to approximate posterior sampling for blind inverse problems, by using multiple score functions designed to estimate each part of the component. With extensive experiments, we show that BlindDPS establishes state-of-the-art performance on both blind deblurring and imaging through turbulence, even when the degradation and the measurement noise are heavy.


{\small
\bibliographystyle{ieee_fullname}
\bibliography{egbib}
}

\clearpage
\appendix

\onecolumn
\renewcommand\thefigure{\thesection.\arabic{figure}}
\renewcommand{\thetable}{A.\arabic{table}}
\counterwithin{figure}{section}
\counterwithin{table}{section}

\section*{\Large{\textbf{Supplementary Material}}}

\section{Proofs}
\label{sec:proofs}
\begin{figure}[t]
\vspace{-0.4cm}
\begin{center}
\begin{tikzpicture}[
Nodex/.style={circle, draw=red!80, fill=red!5, very thick, minimum size=10mm},
Nodek/.style={circle, draw=blue!80, fill=blue!5, very thick, minimum size=10mm},
Nodey/.style={circle, draw=green!80, fill=green!10, very thick, minimum size=10mm}
]
\node[Nodey]  (y) {$\y$}  ;
\node[Nodex]    (x0)     [above right=of y] {$\x_0$} ;
\node[Nodex]  (xt)   [right=of x0] {$\x_t$};
\node[Nodek]    (k0)     [below right=of y] {$\kb_0$};
\node[Nodek]    (kt)     [right=of k0] {$\kb_t$};
\filldraw[black] (1.5,0) circle (0pt) node {$p(\y|\x_0,\kb_0)$};
%
\draw [->, thick] (x0) to [out=-150,in=60]  (y);
\draw [->, thick] (x0) to [out=30,in=150] node[above] {$p(\x_t|\x_0)$} (xt);
\draw [->, thick,dotted] (xt) to [out=210,in=-30] node[below,sloped] {$p(\x_0|\x_t)$} (x0);
\draw [->, thick] (k0) to [out=150,in=-60]  (y);
\draw [->, thick] (k0) to [out=30,in=150] node[above] {$p(\kb_t|\kb_0)$} (kt);
\draw [->, thick,dotted] (kt) to [out=210,in=-30] node[below,sloped] {$p(\kb_0|\kb_t)$} (k0);
\end{tikzpicture}
\captionof{figure}{Probabilistic graph of BlindDPS for blind deblurring.}
\label{fig:prob_graph_bd}
\end{center}
\end{figure}
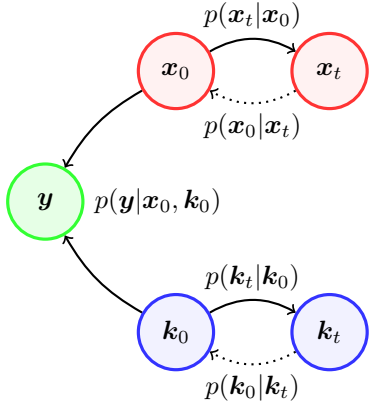

We first borrow the result from~\cite{chung2022diffusion}.
\begin{restatable}{proposition}{tweedie}
\label{prop:post}
For the case of VP-SDE or DDPM sampling whose the forward diffusion is given by
\begin{align}\label{eq:ddpm}
\x_t = \sqrt{{\bar\alpha(t)}}\x_0+\sqrt{1-{\bar\alpha(t)}}\z,\quad\quad \z \sim \Nc(\bm{0}, \Ib),
\end{align}
$p(\x_0|\x_t)$ has the unique posterior mean at
\begin{align}
\label{eq:post}
 \hat \x_0 := \Ed[\x_0|\x_t] &= \frac{1}{\sqrt{{\bar\alpha(t)}}}(\x_t + (1 - {\bar\alpha(t)})\nabla_{\x_t} \log p_t(\x_t)).
\end{align}
\end{restatable}
In our case, Proposition~\ref{prop:post} holds for both the reverse conditional probability $p(\x_0|\x_t)$ as well as $p(\kb_0|\kb_t)$, as they are both constructed from DDPM. Given the posterior mean $\hat\x_0, \hat\kb_0$ that can be computed efficiently (i.e. via one forward pass through the neural network) during the intermediate steps, our proposal is to find a tractable approximation for $p(\y|\x_t, \kb_t)$. Specifically, we propose the following approximation
\begin{align}
\label{eq:hatx0_postmean}
    p(\y|\x_t, \kb_t) \simeq p(\y|\hat\x_0, \hat\kb_0),\quad \mbox{where}\quad &\hat\x_0:=\Ed[\x_0|\x_t]=\Ed_{\x_0 \sim p(\x_0|\x_t)}\left[\x_0\right]\\
    &\hat\kb_0:=\Ed[\kb_0|\kb_t]=\Ed_{\kb_0 \sim p(\kb_0|\kb_t)}\left[\kb_0\right]
\label{eq:hatk0_postmean}
\end{align}
Now, to quantify the approximation error induced by eq.~\eqref{eq:hatx0_postmean},\eqref{eq:hatk0_postmean}, the following definition is useful.
\begin{restatable}[Jensen gap~\cite{gao2017bounds,simic2008global}]{definition}{jensengap}
\label{def:jensengap}
Let $\x$ be a random variable with distribution $p(\x)$. For some function $f$ that may or may not be convex, the Jensen gap is defined as
\begin{align}
     \Jc(f, \x \sim p(\x)) = \Ed[f(\x)] - f(\Ed[\x]),
\end{align}
where the expectation is taken over $p(\x)$.
\end{restatable}
Using the Jensen gap defined in Definition~\ref{def:jensengap}, we attempt to achieve a meaningful upper bound on the gap. First, we have the following
\begin{restatable}[Jensen gap upper bound~\cite{gao2017bounds}]{proposition}{jensengapupperbound}
\label{prop:jensen_gap_upper_bound}
Define the absolute cenetered moment as $m_p := \sqrt[p]{\Ed[|X - \mu|^p]}$, and the mean as $\mu = \Ed[X]$. Assume that for $\alpha > 0$, there exists a positive number $K$ such that for any $x \in \Rd, |f(x) - f(\mu)| \leq K|x - \mu|^\alpha$. Then,
\begin{align}
    |\Ed[f(X) - f(\Ed[X])]| &\leq \int |f(X) - f(\mu)|dp(X) \leq K\int |x - \mu|^\alpha dp(X) \leq Mm_\alpha^\alpha.
\end{align}
\end{restatable}
The following lemmas from~\cite{chung2022diffusion} are also useful.
\begin{restatable}[]{lemma}{lipschitzunivariate}
\label{lem:lipschitz_univariate}
Let $\phi(\cdot)$ be a univariate Gaussian density function with mean $\mu$ and variance $\sigma^2$. There exists a constant $L$ such that $\forall x,y \in \Rd$,
\begin{align}
\label{eq:lipschitz_univariate}
    |\phi(x) - \phi(y)| \leq L|x - y|,
\end{align}
where $L = \frac{1}{\sqrt{2\pi\sigma^2}} \exp{(-\frac{1}{2\sigma^2})}$.
\end{restatable}
\begin{restatable}[]{lemma}{lipschitzmultivariate}
\label{lem:lipschitz_multivariate}
Let $\phi(\cdot)$ be an isotropic multivariate Gaussian density function with mean $\mub$ and variance $\sigma^2 \Ib$. There exists a constant $L$ such that $\forall \x,\y \in \Rd^d$,
\begin{align}
\label{eq:lipschitz_multivariate}
    \|\phi(\x) - \phi(\y)\| \leq L\|\x - \y\|,
\end{align}
where $L = \frac{d}{\sqrt{2\pi\sigma^2}}e^{-1/2\sigma^2}$.
\end{restatable}
\ddlaplace*
\begin{proof}
The proof of the theorem is inspired by and builds upon~\cite{chung2022diffusion}.
We first note that $\x_t, \kb_t\, \forall t \in [0, 1]$ are independent (See Fig.~\ref{fig:prob_graph_bd}). Further, $\y$ and $\x_t$ are conditionally independent on $\x_0$; $\y$ and $\kb_t$ are conditionally independent on $\kb_0$. Then, we have the following factorization
\begin{align}
\label{eq:factorization}
    p(\y|\x_t, \kb_t) &= \int p(\y|\x_0, \kb_0) p(\x_0|\x_t) p(\kb_0|\kb_t)\, d\x_0 d\kb_0\\
    &= \Ed_{\x_0 \sim p(\x_0|\x_t),\, \kb_0 \sim p(\kb_0|\kb_t)}[f(\x_0, \kb_0)],
\end{align}
where $f(\x_0, \kb_0) = h(\kb_0 \ast \x_0)$, with $h(\mub)$ denoting the density function of an isotropic multivariate Gaussian density function with mean $\mub$, and variance $\sigma^2 \Ib$. Our proposal is to use the Jensen approximation
\begin{align}
\label{eq:blindDPS_approx}
    p(\y|\x_t, \kb_t) \simeq p(\y|\Ed[\x_0,\kb_0])= p(\y|\hat\x_0, \hat\kb_0),
\end{align}
where the last equality comes from the independency of $\x_0$ and $\kb_0$.
Now we derive the closed-form upper bound of the Jensen gap.
 For simplicity in exposition, let us define $\Kb_0\x_0 \equiv \kb_0 \ast \x_0, \equiv \Xb_0\kb_0$, where $\Kb_0, \Xb_0$ are block Hankel matrices that represent the convolution operation in matrix multiplication. Further, we denote $\bar{\|\Kb_0\|} := \Ed_{\kb_0 \sim p(\kb_0|\kb_t)}[\|\Kb_0\|]$. Our Jensen gap reads
\begin{align}
    \Jc(f, p(\x_0|\x_t)p(\kb_0|\kb_t)) &= |\Ed_{\x_0, \kb_0}[f(\x_0, \kb_0)] - f(\Ed_{\x_0}[\x_0], \Ed_{\kb_0}[\kb_0])|\\
    &\leq \underbrace{|\Ed_{\kb_0, \x_0}[f(\x_0, \kb_0)] - \Ed_{\kb_0}[f(\Ed_{\x}[\x_0], \kb_0)]|}_{\textcircled{\raisebox{-0.9pt}{1}}} + 
    \underbrace{|\Ed_{\kb_0}[f(\Ed_{\x_0}[\x_0], \kb_0)] - f(\Ed[\x_0], \Ed[\kb_0])|}_{\textcircled{\raisebox{-0.9pt}{2}}},
\end{align}
with
\begin{align}
    \textcircled{\raisebox{-0.9pt}{1}} &= |\Ed_{\kb_0}[\Ed_{\xb_0}[f(\x_0, \kb_0)] - f(\Ed_{\xb_0}[\xb_0], \kb_0)]|\\
    &\stackrel{\text{(a)}}{\leq} \Ed_{\kb_0}\left[\int |h(\kb_0 \ast \x_0) - h(\kb_0 \ast \hat\x_0)|dP(\x_0|\x_t)\right]\\
    &\stackrel{\text{(b)}}{\leq} \Ed_{\kb_0}\left[\frac{d}{\sqrt{2\pi\sigma^2}}e^{-1/2\sigma^2}\int \|\Kb_0\x_0 - \Kb_0\hat\x_0\|dP(\x_0|\x_t)\right]\\
    &\leq \Ed_{\kb_0}\left[\frac{d}{\sqrt{2\pi\sigma^2}}e^{-1/2\sigma^2} \|\Kb_0\| \int \|\x_0 - \hat\x_0\|dP(\x_0|\x_t)\right]\\
    &\stackrel{\text{(c)}}{\leq} \Ed_{\kb_0}\left[\frac{d}{\sqrt{2\pi\sigma^2}}e^{-1/2\sigma^2} \|\Kb_0\| m_{1,\x_0}\right]\\
    &\stackrel{\text{(d)}}{\leq} \frac{d}{\sqrt{2\pi\sigma^2}}e^{-1/2\sigma^2} \|\bar\Kb_0\| m_{1,\x_0},
\end{align}
where (a) is from Proposition.~\ref{prop:jensen_gap_upper_bound}, (b) is from Lemma.~\ref{lem:lipschitz_multivariate}, and (c-d) are from the definitions.
Moreover,
\begin{align}
    \textcircled{\raisebox{-0.9pt}{2}} &\leq \int |h(\hat\kb_0 \ast \hat\x_0) - h(\kb_0 \ast \hat\x_0)|dP(\kb_0|\kb_t)\\
    &\leq \frac{d}{\sqrt{2\pi\sigma^2}}e^{-1/2\sigma^2} \int \|\hat\Xb_0 \kb_0 - \hat\Xb_0 \hat\kb_0\|dP(\kb_0|\kb_t)\\
    &\leq \frac{d}{\sqrt{2\pi\sigma^2}}e^{-1/2\sigma^2} \|\hat\Xb_0\|m_{1,\kb_0}.
\end{align}
Hence
\begin{align}
\label{eq:blindDPS_jensen_gap}
    \Jc(f, p(\x_0|\x_t)p(\kb_0|\kb_t)) \leq \frac{d}{\sqrt{2\pi\sigma^2}}e^{-1/2\sigma^2} \left( \|\bar\Kb_0\|m_{1,\x_0} + \|\hat\Xb_0\|m_{1,\kb_0} \right).
\end{align}
where
\begin{align}
m_{1,\x_0} &:= \int \|\x_0 - \hat\x_0\| dP(\x_0|\x_t)\\
m_{1,\kb_0} &:= \int \|\kb_0 - \hat\kb_0\|dP(\kb_0|\kb_t)
\end{align}
We have derived that the approximation \eqref{eq:blindDPS_approx} has the Jensen gap upper bounded by \eqref{eq:blindDPS_jensen_gap}. Finally, taking the derivative of the log to \eqref{eq:blindDPS_approx}, we have that
\begin{align*}
    \nabla_{\x_t} \log p_t(\y|\x_t, \kb_t)
    &\simeq  \nabla_{\x_t} \log p(\y|\hat\x_0(\x_t), \hat\kb_0(\kb_t)) \\
    \nabla_{\kb_t} \log p_t(\y|\x_t, \kb_t)
    &\simeq  \nabla_{\kb_t} \log p(\y|\hat\x_0(\x_t), \hat\kb_0(\kb_t)).
\end{align*}
Note that the approximation error from the Jensen gap approaches to zero as the noise level $\sigma$ increase sufficiently.
\end{proof}

\clearpage
\twocolumn

\section{BlindDPS}
\subsection{Imaging through turbulence}
\label{sec:supp_blinddps_itt}

\begin{figure}[t]
\vspace{-0.4cm}
\begin{center}
\begin{tikzpicture}[
Nodex/.style={circle, draw=red!80, fill=red!5, very thick, minimum size=10mm},
Nodek/.style={circle, draw=blue!80, fill=blue!5, very thick, minimum size=10mm},
Nodep/.style={circle, draw=yellow!80, fill=yellow!15, very thick, minimum size=10mm},
Nodey/.style={circle, draw=green!80, fill=green!10, very thick, minimum size=10mm}
]
\node[Nodey]  (y) {$\y$}  ;
\node[Nodex]    (x0)     [above right=2of y] {$\x_0$} ;
\node[Nodex]  (xt)   [right=of x0] {$\x_t$};
\node[Nodek]    (k0)     [right=1.1of y] {$\kb_0$};
\node[Nodek]    (kt)     [right=of k0] {$\kb_t$};
\node[Nodep]    (p0)     [below right=2of y] {$\phib_0$};
\node[Nodep]    (pt)     [right=of p0] {$\phib_t$};
\filldraw[black] (-1.75,0) circle (0pt) node {$p(\y|\x_0,\kb_0,\phib_0)$};
%
\draw [->, thick] (x0) to [out=-150,in=60]  (y);
\draw [->, thick] (x0) to [out=30,in=150] node[above] {$p(\x_t|\x_0)$} (xt);
\draw [->, thick,dotted] (xt) to [out=210,in=-30] node[below,sloped] {$p(\x_0|\x_t)$} (x0);
\draw [->, thick] (k0) to [out=180,in=0]  (y);
\draw [->, thick] (k0) to [out=30,in=150] node[above] {$p(\kb_t|\kb_0)$} (kt);
\draw [->, thick,dotted] (kt) to [out=210,in=-30] node[below,sloped] {$p(\kb_0|\kb_t)$} (k0);
\draw [->, thick] (p0) to [out=150,in=-60]  (y);
\draw [->, thick] (p0) to [out=30,in=150] node[above] {$p(\phib_t|\phi_0)$} (pt);
\draw [->, thick,dotted] (pt) to [out=210,in=-30] node[below,sloped] {$p(\phib_0|\phib_t)$} (p0);
\end{tikzpicture}
\captionof{figure}{Probabilistic graph of BlindDPS for imaging through turbulence.}
\label{fig:prob_graph_t}
\end{center}
\end{figure}
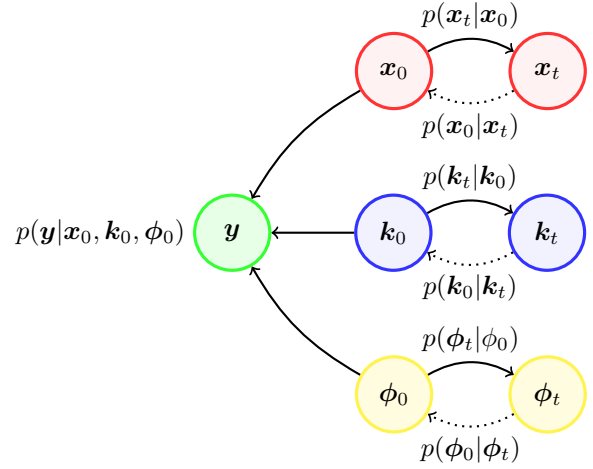

In terms of inverse problem solving, the tilt-blur model is often used~\cite{chan2022tilt,shimizu2008super,chak2018subsampled}, as the model is simple but fairly accurate. Specifically, we have
\begin{align}
\label{eq:forward_turbulence}
    p(\y|\x_0, \kb_0, \bm{\phi}_0) := \Nc(\y|\kb_0 \ast \Tc_{\bm{\phi}_0}(\x_0), \sigma^2 \Ib) .
\end{align}
For details in the forward model that is used for our experiments, see Supplementary Section~\ref{sec:ip_setting}. 
Note that the three factors are all independent, i.e.
\begin{align*}
    p(\x_0, \kb_0, \bm{\phi}_0|\y) \propto p(\y|\x_0, \kb_0, \bm{\phi}_0)p(\x_0)p(\kb_0)p(\bm{\phi}_0).
\end{align*}
Then, from Remark~\ref{rmk:generalize}, we can again construct a system of reverse SDEs (See Fig.~\ref{fig:prob_graph_t}) analogous to the blind deblurring case (\eqref{eq:reverse-sde-x},\eqref{eq:reverse-sde-k}):
\begin{align}
    d\x &= (-\frac{\beta(t)}{2}\x - \beta(t)[\nabla_{\x_t} \log p(\y|\hat\x_0(\x_t), \hat\kb_0(\kb_t), \hat{\bm{\phi}}_0(\bm{\phi}_t))\notag \\&+ \s_{\theta^*}^i(\x_t,t)] )dt + \sqrt{\beta(t)}d\bar\w,\\
    d\kb &= (-\frac{\beta(t)}{2}\kb - \beta(t)[\nabla_{\kb_t} \log p(\y|\hat\x_0(\x_t), \hat\kb_0(\kb_t), \hat{\bm{\phi}}_0(\bm{\phi}_t))\notag \\&+ \s_{\theta^*}^k(\kb_t,t)] )dt + \sqrt{\beta(t)}d\bar\w,\\
    d\bm{\phi} &= (-\frac{\beta(t)}{2}\kb - \beta(t)[\nabla_{\bm{\phi}_t} \log p(\y|\hat\x_0(\x_t), \hat\kb_0(\kb_t), \hat{\bm{\phi}}_0(\bm{\phi}_t))\notag \\&+ \s_{\theta^*}^t(\bm{\phi}_t,t)] )dt + \sqrt{\beta(t)}d\bar\w,
\end{align}
where $\s_{\theta^*}^t$ is the score function trained to model the distribution of the tilt maps. Then, we can construct a similar method as shown in Algorithm~\ref{alg:itt} based on ancestral sampling analogous to Algorithm~\ref{alg:db}. Note that for solving imaging through turbulence, we do not use the $\ell_0$ sparsity prior.

\begin{algorithm}[!t]
\caption{BlindDPS --- Imaging through turbulence}
\begin{algorithmic}[1]
\Require $N$, $\y$, $\alpha,  {\{\tilde\sigma_i\}_{i=1}^N}$
\State $\x_N,\kb_N,\phib_N \sim \Nc(\bm{0}, \bm{I})$
\For{$i=N-1$ {\bfseries to} $0$}
 \State{{$\hat\s^i \gets \s_{\theta^*}^i(\x_i, i)$}}
 \State{{$\hat\s^k \gets \s_{\theta^*}^k(\kb_i, i)$}}
 \State{{$\hat\s^k \gets \s_{\theta^*}^t(\phib_i, i)$}}
 \State{{$\hat\x_0 \gets \frac{1}{\sqrt{\bar\alpha_i}}(\x_i + \sqrt{1 - \bar\alpha_i}\hat\s^i)$}}
 \State{{$\hat\kb_0 \gets \frac{1}{\sqrt{\bar\alpha_i}}(\kb_i + \sqrt{1 - \bar\alpha_i}\hat\s^k)$}}
 \State{{$\hat\kb_0 \gets \Pc_C(\hat\kb_0)$}}
 \State{{$\hat\phib_0 \gets \frac{1}{\sqrt{\bar\alpha_i}}(\phib_i + \sqrt{1 - \bar\alpha_i}\hat\s^t)$}}
 \State{$\z_i,\z_k,\z_t \sim \Nc(\bm{0}, \bm{I})$}
 \State{$\x'_{i-1} \gets \frac{\sqrt{\alpha_i}(1-\bar\alpha_{i-1})}{1 - \bar\alpha_i}\x_i + \frac{\sqrt{\bar\alpha_{i-1}}\beta_i}{1 - \bar\alpha_i}\hat\x_0 +  {\tilde\sigma_i \z_i}$}
 \State{$\kb'_{i-1} \gets \frac{\sqrt{\alpha_i}(1-\bar\alpha_{i-1})}{1 - \bar\alpha_i}\kb_i + \frac{\sqrt{\bar\alpha_{i-1}}\beta_i}{1 - \bar\alpha_i}\hat\kb_0 +  {\tilde\sigma_i \z_k}$}
 \State{$\phib'_{i-1} \gets \frac{\sqrt{\alpha_i}(1-\bar\alpha_{i-1})}{1 - \bar\alpha_i}\phib_i + \frac{\sqrt{\bar\alpha_{i-1}}\beta_i}{1 - \bar\alpha_i}\hat\phib_0 +  {\tilde\sigma_i \z_t}$}
 \State{$\x_{i-1} \gets \x'_{i-1} - \alpha\nabla_{\x_i}\|\y - \hat\kb_0 \ast \Tc_{\phib_0}(\hat\xb_0)\|_2$}
 \State{$\kb_{i-1} \gets \kb'_{i-1} - \alpha\nabla_{\kb_i}\|\y - \hat\kb_0 \ast \Tc_{\phib_0}(\hat\xb_0)\|_2$}
 \State{$\phib_{i-1} \gets \phib'_{i-1} - \alpha\nabla_{\phib_i}\|\y - \hat\kb_0 \ast \Tc_{\phib_0}(\hat\xb_0)\|_2$}
\EndFor
\State {\bfseries return} ${\x}_0, \kb_0, \phib_0$
\end{algorithmic}\label{alg:itt}
\end{algorithm}

\section{Detailed Ablation Studies}
\label{sec:detailed_ablation_studies}

\begin{algorithm}[!t]
\caption{Diffusion Posterior Sampling --- Uniform prior}
\begin{algorithmic}[1]
\Require $N$, $\y$, $\alpha_\x, \alpha_\kb {\{\tilde\sigma_i\}_{i=1}^N}, \lambda, \sigma_{\rm init}$
\State $\x_N \sim \Nc(\bm{0}, \bm{I})$
\State $\kb_N \sim {\rm GaussianKernel(\sigma_{\rm init})}$
\For{$i=N-1$ {\bfseries to} $0$}
 \State{{$\hat\s^i \gets \s_{\theta^*}^i(\x_i, i)$}}
 \State{{$\hat\x_0 \gets \frac{1}{\sqrt{\bar\alpha_i}}(\x_i + \sqrt{1 - \bar\alpha_i}\hat\s^i)$}}
 \State{{$\kb_i \gets \Pc_C(\kb_i)$}}
 \State{$\z_i \sim \Nc(\bm{0}, \bm{I})$}
 \State{$\x'_{i-1} \gets \frac{\sqrt{\alpha_i}(1-\bar\alpha_{i-1})}{1 - \bar\alpha_i}\x_i + \frac{\sqrt{\bar\alpha_{i-1}}\beta_i}{1 - \bar\alpha_i}\hat\x_0 +  {\tilde\sigma_i \z_i}$}
 \State{$\x_{i-1} \gets \x'_{i-1} - \alpha_{\x}\nabla_{\x_i}\|\y - \kb_i \ast \hat\xb_0\|_2$}
 \State{$\Lc_\kb \gets \|\y - \kb_i \ast \hat\xb_0\|_2 + \lambda\ell_0(\kb_i)$}
 \State{$\kb_{i-1} \gets \kb_{i} - \alpha_{\kb}\nabla_{\kb_i}\Lc_\kb$}
\EndFor
\State {\bfseries return} $\x_0, \kb_0$
\end{algorithmic}\label{alg:db_uniform_prior}
\end{algorithm}

\subsection{Diffusion prior for the forward model}

Let us revisit the Bayes' rule in the context of diffusion models for posterior sampling in blind deconvlution:
\begin{align*}
    \nabla_{\x_t} \log p(\x_t, \kb_t|\y) &= \nabla_{\x_t} \log p(\y|\x_t, \kb_t) + \nabla_{\x_t} \log p(\x_t) \\
    \nabla_{\kb_t} \log p(\x_t, \kb_t|\y) &= \nabla_{\kb_t} \log p(\y|\x_t, \kb_t) + \nabla_{\kb_t} \log p(\kb_t).
\end{align*}
We consider the case where we construct the diffusion prior for the image $\x$, but not for the kernel $\kb$. In fact, this setting is similar to the concurrent work of Levac {\em et al.}~\cite{levac2022accelerated}, where the authors propose to use a score function only for the image, and not for the parameters for the motion artifact generating forward model. Note that the {\em blind} forward model setting here is considerably simpler than our method, since the parameter $\kappa$ to be estimated is a scalar. In this regard, the authors propose to use a {\em uniform} prior for the unknown parameter $\kappa$, which makes the gradient of the prior to be simply 0, i.e. $\nabla_{\kappa_t} \log p(\kappa_t) = 0$. If we apply such uniform prior to our setting, our discretized update rule reads
\begin{align*}
    \nabla_{\x_i} \log p(\x_i, \kb_i|\y) &\simeq \s_{\theta^*}^i(\x_i,i) - \frac{1}{\sigma^2} \nabla_{\x_i} \|\y - \kb_i \ast \hat\x_0(\x_i)\|_2^2\\
    \nabla_{\kb_i} \log p(\x_i, \kb_i|\y) &\simeq - \frac{1}{\sigma^2} \nabla_{\kb_i} \|\y - \kb_i \ast \hat\x_0(\x_i)\|_2^2.
\end{align*}
Additionally, similar to BlindDPS, one can further augment sparsity to the kernel estimation by using e.g. $\ell_0$ regularization. Combined with the ancestral sampling steps, we arrive at Algorithm~\ref{alg:db_uniform_prior}. Note that we chose Gaussian kernel as an initialization, but other choices are also feasible.
The main difference between BlindDPS (Algorithm~\ref{alg:db}) and Algorithm~\ref{alg:db_uniform_prior} comes from the the complexity of the priors used. In order to quantify the performance gap, we chose 100 images from the FFHQ validation set, and compared the result of Algorithm~\ref{alg:db_uniform_prior} against BlindDPS. We performed grid search to find the optimal parameters $\alpha_\x, \alpha_\kb, \lambda$, which were set to $\alpha_\x = 0.3$, $\alpha_\kb = 0.3$, and $\lambda = 5.0$.

\begin{table}[t]
\centering
\setlength{\tabcolsep}{0.2em}
\resizebox{0.45\textwidth}{!}{
\begin{tabular}{llll@{\hskip 15pt}lll}
\toprule
{} & \multicolumn{3}{c}{\textbf{Image}} & \multicolumn{2}{c}{\textbf{Kernel}} \\
\cmidrule(lr){2-4}
\cmidrule(lr){5-6}
{\textbf{Method}} & {LPIPS $\downarrow$} & {PSNR $\uparrow$} & {SSIM $\uparrow$} & {MSE $\downarrow$} & {MNC $\uparrow$}\\
\midrule
BlindDPS~\textcolor{trolleygrey}{(ours)} & \textbf{0.247} & \textbf{23.65} & \underline{0.786} & \textbf{0.002} & \textbf{0.958}\\
Uniform prior~\cite{levac2022accelerated} & 0.566 & 11.72 & 0.369 & 0.163 & 0.844\\
\bottomrule
\end{tabular}
}
\vspace{0.2em}
\caption{
Ablation study: uniform prior vs. diffusion prior (BlindDPS).
}
\label{tab:ablation_uniform_prior_quantitative}
\end{table}

\begin{figure*}
     \centering
     \begin{subfigure}[b]{0.49\textwidth}
         \centering
         \includegraphics[width=\textwidth]{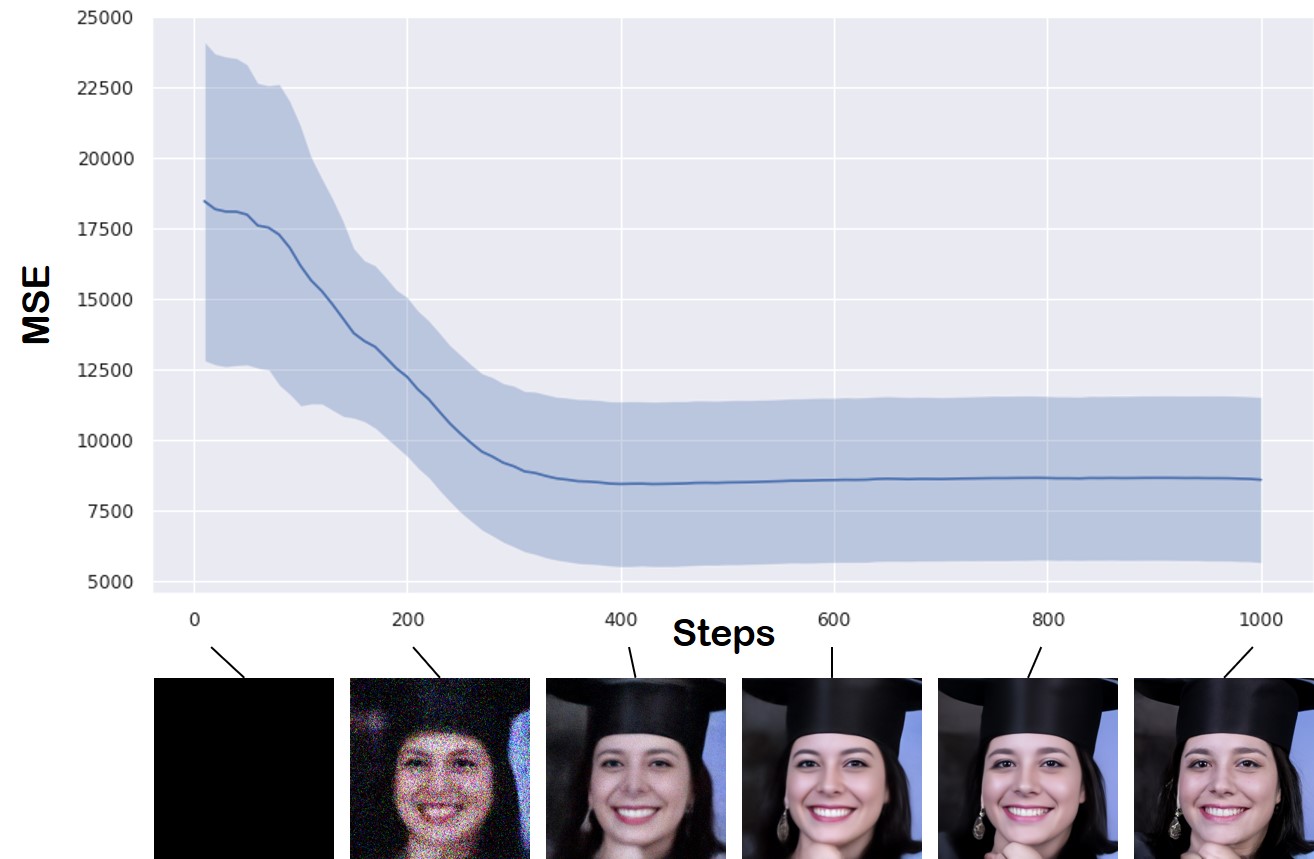}
         \caption{Progress of $\hat\x_0(\x_t)$}
         \label{fig:progress_of_x}
     \end{subfigure}
     \begin{subfigure}[b]{0.49\textwidth}
         \centering
         \includegraphics[width=\textwidth]{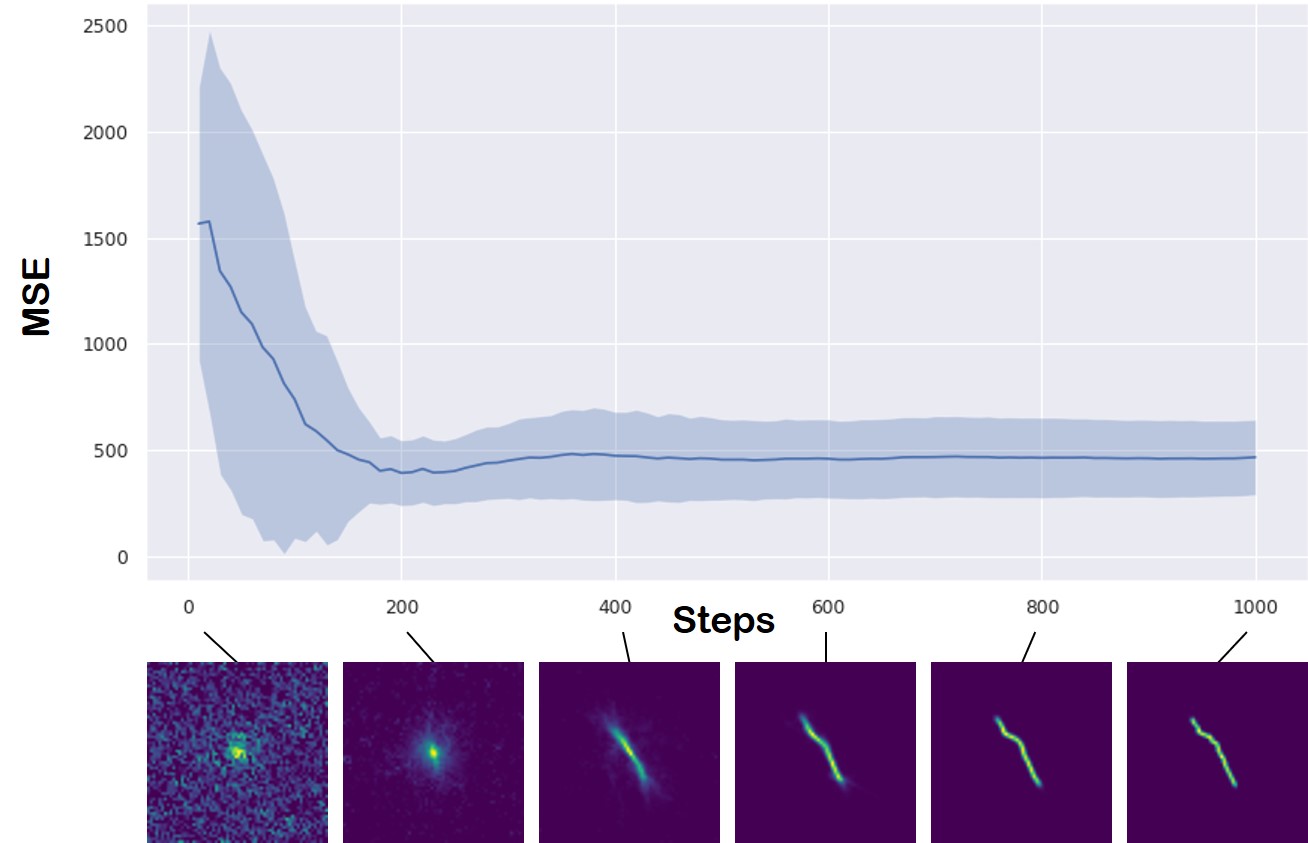}
         \caption{Progress of $\hat\kb_0(\kb_t)$}
         \label{fig:progress_of_k}
     \end{subfigure}
    \label{fig:trend}
    \caption{Progress of estimation error averaged over 100 test set in blind deconvolution. Blue line: mean value, shaded area: $\pm 1\sigma$. Measured with MSE against the ground truth.}
\end{figure*}

Representative results can be seen in Fig.~\ref{fig:ablation_uniformprior}, and quantitative results can be found in Table~\ref{tab:ablation_uniform_prior_quantitative}. Clearly, uniform prior far underperforms against the diffusion prior proposed in this work. We can conclude that while simple priors such as uniform prior may be a feasible option for {\em scalar} parameters, as the one in~\cite{levac2022accelerated}, much care should be taken when applied to higher dimensional parameters such as blind deconvolution.

\subsection{{Effect of sparsity regularization}}
{To check the effect of sparsity regularization in \eqref{eq:sparsity-reg}, we perform an ablation study by varying $\lambda$ from $0.0$ to $5.0$.
Specifically, we use $l_1$ sparsity regularization with different $\lambda$ for 100 blurred images taken from validation set for FFHQ, with forward model and blur kernels adjusted to be identical to those of the main experiment (section~\ref{sec:ip_setting}).}

\subsection{Progress of estimation}

As discussed in section~\ref{sec:blindDPS} of main text, the proposed method admits a natural Gaussian scale-space evolution of estimation, when visualized in the {\em denoised} representations $\hat\x_0, \hat\kb_0$. To quantify the trend in which the estimates evolve, we measure the MSE against the ground truth image and the kernel, and average the trend over 100 of the test data. We summarize the result in Fig.~\ref{fig:progress_of_x}, \ref{fig:progress_of_k}. Here, we see that the MSE value drops to the minimum value at about 400/1000, 200/1000 iterations, which is relatively early in the whole reverse diffusion process. For the rest of the steps (especially for the images), the remaining high frequency details are in-filled, boosting the perceptual quality.

\section{Extended Related Works}
\label{sec:extended_related_works}

In this section, we discuss related works categorized into two applications that we tackle - blind deblurring, and imaging through turbulence.

\subsection{Blind deblurring}

We first review the optimization-based (model-based) methods that were extensively studied. 
The seminal work of Chan {\em et al.}~\cite{chan1998total} introduced the total variation  (TV) prior, which enhances the gradient sparsity of both the image and the kernel. The scheme has been developed and re-invented over the years~\cite{levin2009understanding}, yielding better practices to obtain stable results~\cite{perrone2014total}. To promote sparsity of both the image and the kernel, regularizations based on $\ell_0$ penalty~\cite{pan2016l_0}, $\ell_p,\, 0 < p < 1$ penalty~\cite{zuo2016learning} based on the generalized iterative shrinkage algorithm (GISA)~\cite{zuo2013generalized}, $\ell_1, \ell_2$~\cite{krishnan2011blind} were proposed. Later on, it was shown that non-blurry natural images have sparse ``dark channel''~\cite{Pan_2016_CVPR}, where the dark channel is computed as the union of minimum values in patch occurrences. Promoting sparsity of the dark channel~\cite{pan2017deblurring} has shown to be an effective method for performing blind deconvolution. When the regularization functions are chosen, one typically performs alternating optimization strategies~\cite{boyd2011distributed} to solve the problem. It should be noted that it is often the case where the optimization strategy is non-trivial, and involves many tricks such as multi-scale optimization~\cite{pan2016blind}, and painful parameter tuning for specific input images. Wrong choice of parameter/optimization strategy typically results in heavily compromised performance.

In recent years, deep learning (DL) based methods have been largely developed. One can categorize DL methods into 1) explicit kernel estimation methods, where the network is designed to both deblur the image, and to estimate the exact kernel; 2) amortized inference, where the estimation of kernel does not take place. For the first type of methods, convolutional neural networks (CNN) were adopted for seperate modules, estimating the kernel and the deblurred image, respectively~\cite{schuler2015learning,sun2015learning,xu2017motion}. Advancing the conventional model-based priors, discriminative priors~\cite{li2018learning} and deep image priors (DIP)~\cite{ren2020neural} were proposed, showing improved performance. While deep priors typically improves the performance, one should note that they are also often unstable, leading to undesirable solutions: both adversarial training and jointly training two deep image priors are hard to handle.

More recently, learning the inverse mapping without explicitly estimating the kernel has gained popularity. For these methods, neural network is trained through supervised learning with paired clean and blurry images. Especially, DeblurGAN~\cite{kupyn2018deblurgan} used the perceptual loss that helps to maintain contents and adversarial loss that minimizes the Wasserstein distance between the clean images and reconstructed images. DeblurGAN-v2~\cite{kupyn2019deblurgan} focused on handling multi-scale features to solve the blind deblurring problem. They adopted Feature Pyramide Network (FPN) and proposed double-scale discriminators, where each discriminator measures the Wasserstein distance between clean images and reconstructed images at global and local patch level, respectively. Meanwhile, MPRNet~\cite{zamir2021multi} adopted a multi-stage learning method that decomposes the given problem into sub-problems and solves each one through a lightweight sub-network including a supervised attention module that gives weight to local features. As a result, blurry images are progressively restored. On the other hand, transformer based methods has been proposed and shown notable performance on deblurring task. Specifically, IPT~\cite{chen2021pre} pretrained transformer on multiple image processing tasks and fine-tune the transformer on each tasks, Uformer~\cite{wang2022uformer} proposed LeWin transformer block for locally-enhanced self attention and multi-scale modulator, and Restormer~\cite{zamir2022restormer} proposed two specialized transformer modules called MDTA and GDFN with progressive training scheme that enhances the image restoration performance on different spatial resolutions.
While often achieving state-of-the-art performance, these methods tend to compromise flexibility, modularity, and generalization capacity. For instance, the model cannot handle degradations that deviate from the traning data.

\subsection{Imaging through turbulence}

Although the correct estimation model for imaging through turbulence  is tilt-then-blur~\cite{chan2022tilt}, for inverse problem solving, the blur-then-tilt model is more often used. This is mainly due to the ease of applying off-the-shelf blind deblurring methods once the tilt is mitigated through, e.g. optical flow~\cite{liu2009beyond}. While in our work, we only consider single frame turbulence mitigation for simplicity, it is usually the case where we have multiple temporal frames that are degraded by random phase distortions. Hence, removing the tilt proceeds by e.g. temporal averaging~\cite{zhu2012removing}, variational model~\cite{xie2016removing}, frame selection~\cite{anantrasirichai2013atmospheric}, etc. Moreover, when dealing with sequence of images, the ``Lucky image fusion'' step is often performed to find the reference image with the least amount of phase distortion. For details in such step, see, e.g.~\cite{fried1978probability}.
Once the distortion (tilt) is mitigated, the deblurring step is often performed with off-the-shelf algorithms~\cite{zhu2012removing,anantrasirichai2013atmospheric,xie2016removing}. However, as most off-the-shelf deblurring algorithms do not take into account the kernel priors specifically for turbulence, a more specified algorithm leveraging basis expansion~\cite{mao2020image} was proposed.

Similar to deblurring methods, various DL based methods have been proposed. Utilizing CNN to estimate the phase distortion map~\cite{liu2019deep} was proposed. Moreover, supervised learning based on pairs of simulated atmospheric turbulence images have been proposed over the years. Transfer learning approach from pre-trained deblurring network was proposed~\cite{guo2022blind}. Variants of generative adversarial network (GAN) based methods were also proposed~\cite{jin2021neutralizing,rai2022removing}, leveraging the adversarial learning scheme to enhance the visual quality of the reconstructions. Recently, a method that uses physics-driven transformer architecture dubbed TurbNet~\cite{mao2022single} was proposed. To the best of our knowledge, none of the methods in the literature considered using unsupervised reconstruction scheme by utilizing the generative prior, as in our method. Although our method is developed upon a rather simplified forward model of imaging through turbulence, we believe our work establishes a proof of concept, and opens up a new are regarding turbulence reconstruction.

\section{Inverse problem setting}
\label{sec:ip_setting}

In this section, we briefly summarize how our forward model is constructed.

\subsection{Blind deblurring}
The forward model is given as
\begin{align}
    \y = \kb_0 \ast \xb_0 + \nb,\,\nb \sim \Nc(0, \sigma^2\Ib),
\end{align}
where $\sigma = 0.02$ is set as the measurement noise level. The size of the kernel is set to $64\times64$. For motion blur kernels, we use the random kernel generator from\footnote{\url{https://github.com/LeviBorodenko/motionblur/blob/master/motionblur.py}} with intensity value set to 0.5.

\subsection{Imaging through turbulence}
The forward model is given as
\begin{align}
    \y = \kb_0 \ast \Tc_{\bm{\phi}_0}(\x_0) + \nb,\, \nb \sim \Nc(0, \sigma^2\Ib),
\end{align}
where $\bm{\phi}$ is the tilt vector field that has identical size of the given image (i.e. in our case $256\times 256$). Specifically, the tilt vector field is generated with the algorithm proposed in~\cite{chak2018subsampled}. The parameters are set to $M = 500, N = 32, \sigma = 1.0$, with all the other parameters set same to the baseline. The blur kernel $\kb_0$ is taken to be isotropic Gaussian kernel with standard deviation of 0.4 (FFHQ), and 0.2 (ImageNet). The proposed algorithm for solving imaging through turbulence is presented in Algorithm.~\ref{alg:itt}.

\section{Experimental Details}
\label{sec:exp_details}

\subsection{Training}
We take pre-trained score function for the FFHQ dataset, and the ImageNet dataset, following the settings of~\cite{chung2022diffusion}. When training the score function for kernels, we create a database of that consists of 60k $64\times64$ kernels. Among them, 50k motion blur kernels were generated from\footnote{\url{https://github.com/LeviBorodenko/motionblur}}, by sampling the intensity value $I \sim {\rm Unif}(0.2, 1.0)$. The other 10k Gaussian blur kernels were generated with random standard deviation $\sigma \sim {\rm Unif}(0.1, 5.0)$.

For training the score function for kernel / tilt-map, we use the U-Net architecture from \code{guided-diffusion}\footnote{https://github.com/openai/guided-diffusion}, and train the models using base configurations. The models were trained with a single RTX 3090 GPU for 3.0M / 1.5M steps, which took about one day / two days, respectively.

\subsection{Compute time}
As stated in the limitations, the number of score functions that are used at inference time scales linearly with the number of components involved in the forward model. For blind deblurring, two neural networks are used (image, kernel), and for imaging through turbulence, three neural networks are used (image, kernel, tilt map). In order to quantify additional compute cost in each of the situation, we measure the wall-clock time to reconstruct a single image with a single RTX 2080ti GPU. 
DPS~\cite{chung2022diffusion}: 132.39 sec. BlindDPS---Blind deblurring(2 score functions): 180.22 sec. BlindDPS---Imaging through turbulence(3 score functions): 220.76 sec.

\subsection{Comparison methods}

\noindent
\textbf{Pan-DCP~\cite{pan2017deblurring}.~}
The method utilizes the dark channel prior as the regularization function for images. We use the official implementation\footnote{\url{https://jspan.github.io/projects/dark-channel-deblur/index.html}}, with the parameters advised for facial blur images. We list the specific parameters below. Optimization is performed in a coarse-to-fine strategy in 8 different stages.
\begin{itemize}
    \item $\lambda_{\rm dark} = 4e-3$
    \item $\lambda_{\rm grad} = 4e-3$
    \item $\lambda_{\rm tv} = 1e-3$
    \item $\lambda_{\rm l0} = 5e-4$
\end{itemize}

\noindent
\textbf{Pan-$\ell_0$~\cite{pan2016l_0}.~}
The method regularizes $\ell_0$ regularization for both the image and the kernel. We use the official implementation\footnote{\url{https://jspan.github.io/projects/text-deblurring/index.html}}, with the parameters set as below. Optimization and post-processing is performed similar to Pan-DCP.
\begin{itemize}
    \item $\lambda_{\rm pixel} = 4e-3$
    \item $\lambda_{\rm grad} = 4e-3$
    \item $\lambda_{\rm tv} = 1e-3$
    \item $\lambda_{\rm l0} = 2e-3$
\end{itemize}

\noindent
\textbf{SelfDeblur~\cite{ren2020neural}.~}
We use the default setting of YCbCr deblurring that selfdeblur uses, with static learning rate of $0.01$ for 2500 steps. Optimization is performed by minimizing the MSE for the first 500 steps, and then switching the loss to $1 - SSIM(\cdot, \cdot)$.

\noindent
\textbf{MPRNet~\cite{zamir2021multi}.~}
We use the official implementation\footnote{\url{https://github.com/swz30/MPRNet}}, with the parameters, learning rate decay and neural network architectures advised for the deblurring task. For both FFHQ and AFHQ, we train the model for 30k iterations with a batch size of 3. 
For a fair comparison with the proposed method, half of the input image consists of gaussian blurred images and the other half image consists of motion blurred image.

\noindent
\textbf{DeblurGANv2~\cite{kupyn2019deblurgan}.~}
We use the official implementation\footnote{\url{https://github.com/VITA-Group/DeblurGANv2}}, by following the default settings for parameters, data augmentation strategies and neural network architectures. Specifically, we train the model by minimizing sum of pixel distance loss, WGAN-gp adversarial loss and perceptual loss with weight parameters as below. Inception-ResNet-v2 is used as backbone of the generator. For both FFHQ and AFHQ, we train the model for 1.5 million iterations with a batch size of 1 and input image contains half Gaussian blurred images and the other half motion blurred images for fair comparison with the proposed method.
\begin{itemize}
    \item $\lambda_{\rm pixel} = 5e-1$
    \item $\lambda_{\rm adv} = 6e-3$
    \item $\lambda_{\rm perceptual} = 1e-2$
\end{itemize}

\noindent
\textbf{ILVR~\cite{choi2021ilvr}.~}
We choose the following hyper-paremters: down-scaling factor of 16, 1000 sampling steps, with the latent guidance applied for 1000-100 sampling steps. We use the same score functions that were used for BlindDPS.

\noindent
\textbf{TSR-WGAN~\cite{jin2021neutralizing}.~}
The original work considers spatio-temporal 3D data, whereas our inverse problem setting considers single frame imaging through turbulence. Hence, we design a U-Net like network architecture that consists of 2D convolutions rather than leveraging 3D convolutions. Other training configurations follow the default setting of~\cite{jin2021neutralizing}.

Note that for methods that are capable of estimating the kernel simultaneously (i.e. Pan-DCP, Pan-$\ell_0$, SelfDeblur), only odd-sized kernels can be estimated, whereas our ground truth kernels are even-sized. To match the discrepancy, we estimate 65$\times$65 sized kernel first, and then cut the redundant row/column as the post-processing step. In practice, such discrepancy only affects the result marginally.

\subsection{Data and Code availability}
Our open source implementation will be made public upon publication.

\section{Further Experiments}

\begin{figure*}[t]
  \centering
    \centerline{{\includegraphics[width=1.0\linewidth]{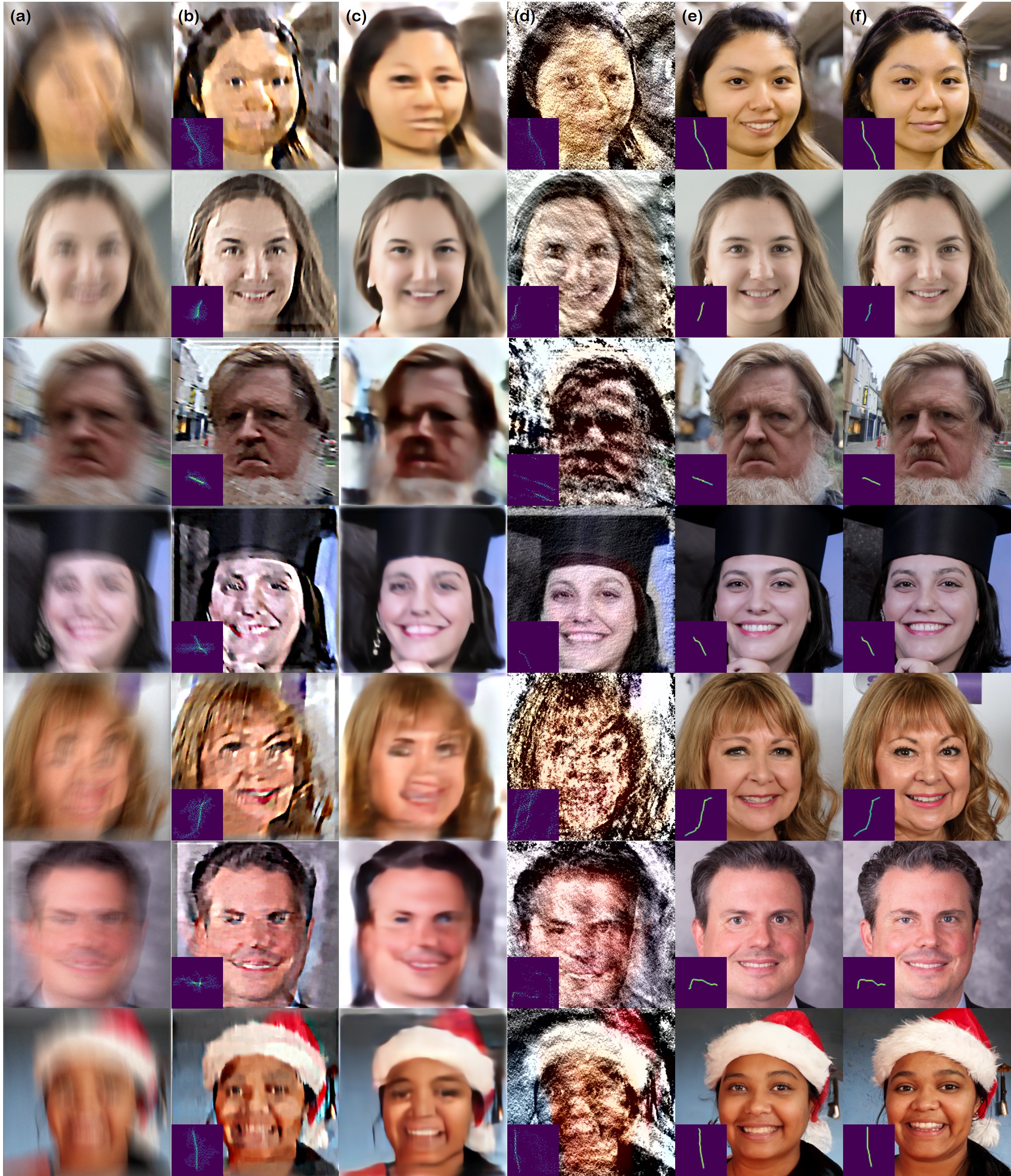}}}
  \caption{Blind motion deblurring results on the FFHQ $256\times 256$ dataset. (a) Measurement, (b) Pan-DCP~\cite{pan2017deblurring}, (c) MPRNet~\cite{zamir2021multi}, (d) SelfDeblur~\cite{ren2020neural}, (e) BlindDPS (\textbf{ours}), (f) Ground truth.}
  \label{fig:results_motion_deblur_ffhq_extensive}
\end{figure*}

\begin{figure*}[t]
  \centering
    \centerline{{\includegraphics[width=1.0\linewidth]{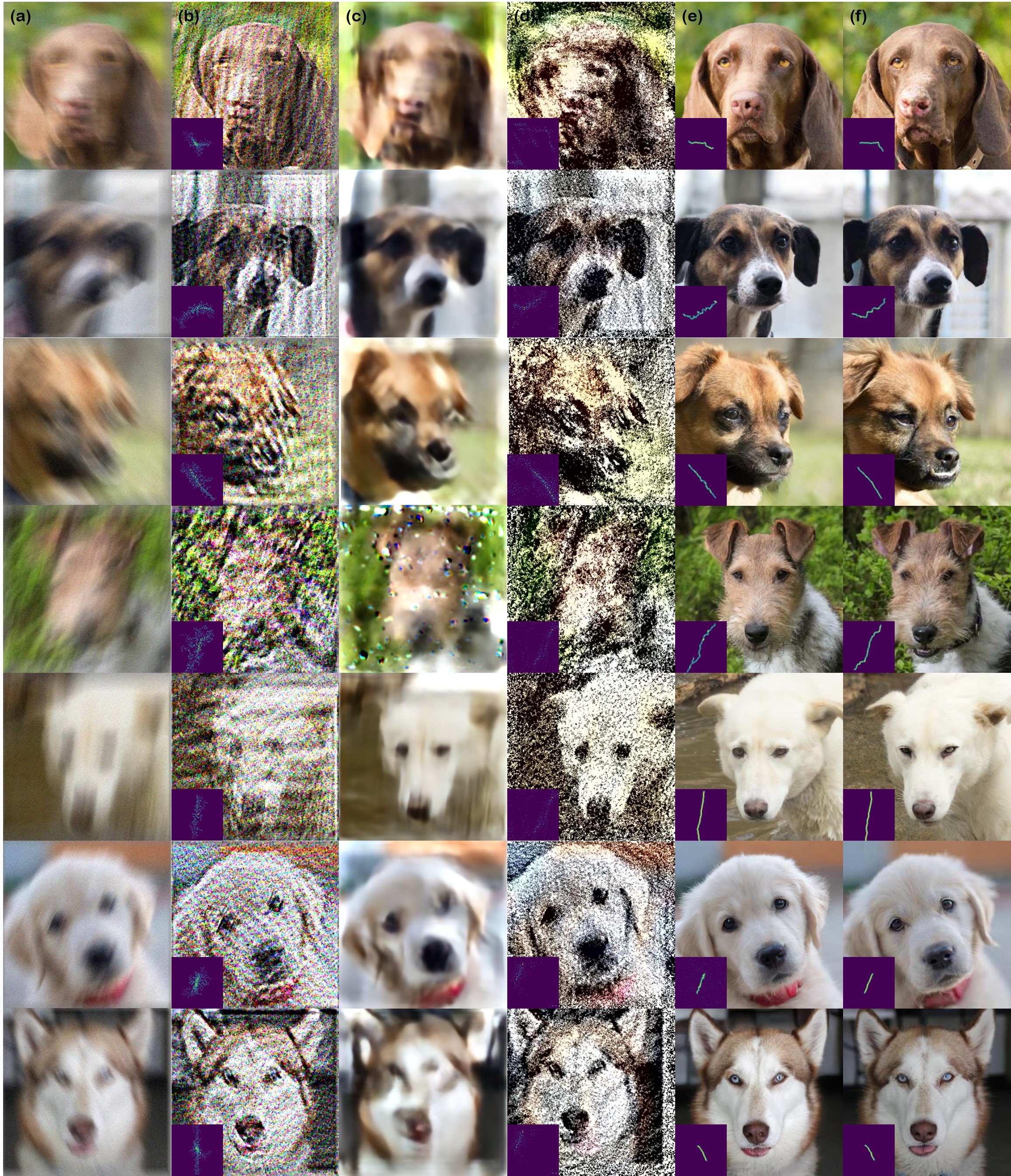}}}
  \caption{Blind motion deblurring results on the AFHQ $256\times 256$ dataset. (a) Measurement, (b) Pan-DCP~\cite{pan2017deblurring}, (c) MPRNet~\cite{zamir2021multi}, (d) SelfDeblur~\cite{ren2020neural}, (e) BlindDPS (\textbf{ours}), (f) Ground truth.}
  \label{fig:results_motion_deblur_afhq_extensive}
\end{figure*}

\begin{figure*}[t]
  \centering
    \centerline{{\includegraphics[width=1.0\linewidth]{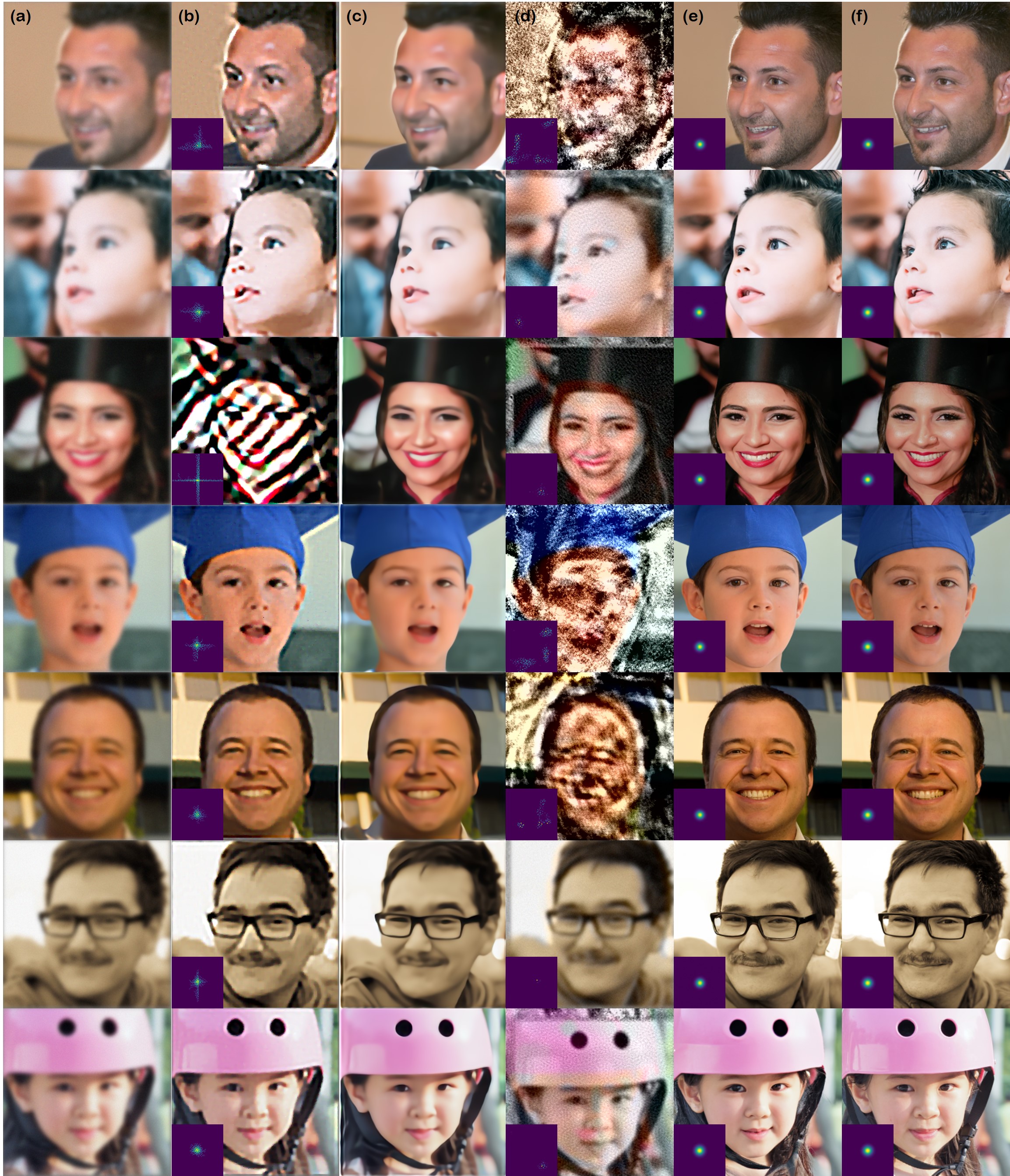}}}
  \caption{Blind Gaussian deblurring results on the FFHQ $256\times 256$ dataset. (a) Measurement, (b) Pan-DCP~\cite{pan2017deblurring}, (c) MPRNet~\cite{zamir2021multi}, (d) SelfDeblur~\cite{ren2020neural}, (e) BlindDPS (\textbf{ours}), (f) Ground truth.}
  \label{fig:results_gaussian_deblur_ffhq_extensive}
\end{figure*}

\begin{figure*}[t]
  \centering
    \centerline{{\includegraphics[width=1.0\linewidth]{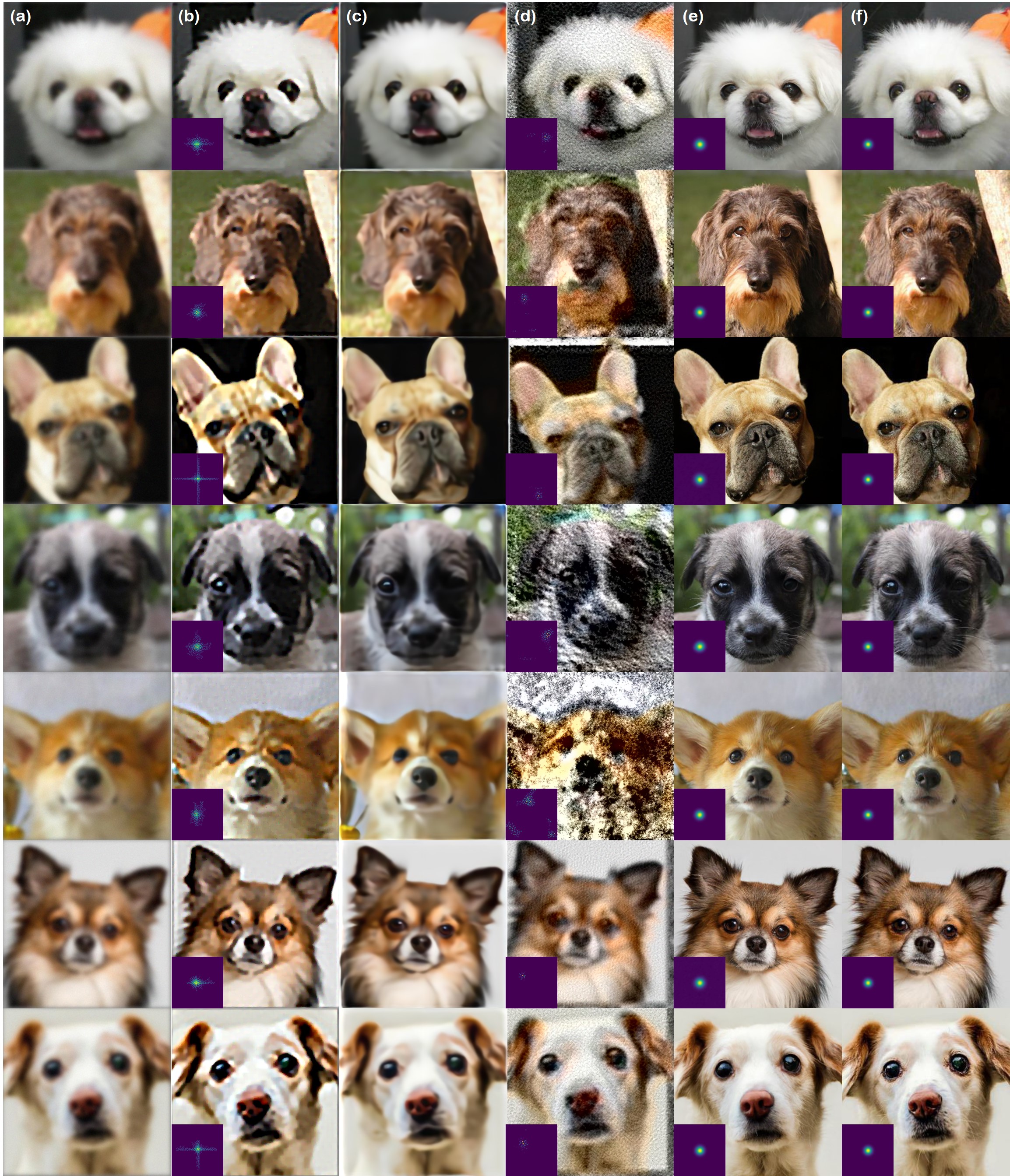}}}
  \caption{Blind Gaussian deblurring results on the AFHQ $256\times 256$ dataset. (a) Measurement, (b) Pan-DCP~\cite{pan2017deblurring}, (c) MPRNet~\cite{zamir2021multi}, (d) SelfDeblur~\cite{ren2020neural}, (e) BlindDPS (\textbf{ours}), (f) Ground truth.}
  \label{fig:results_gaussian_deblur_afhq_extensive}
\end{figure*}

\begin{figure*}[t]
  \centering
    \centerline{{\includegraphics[width=1.0\linewidth]{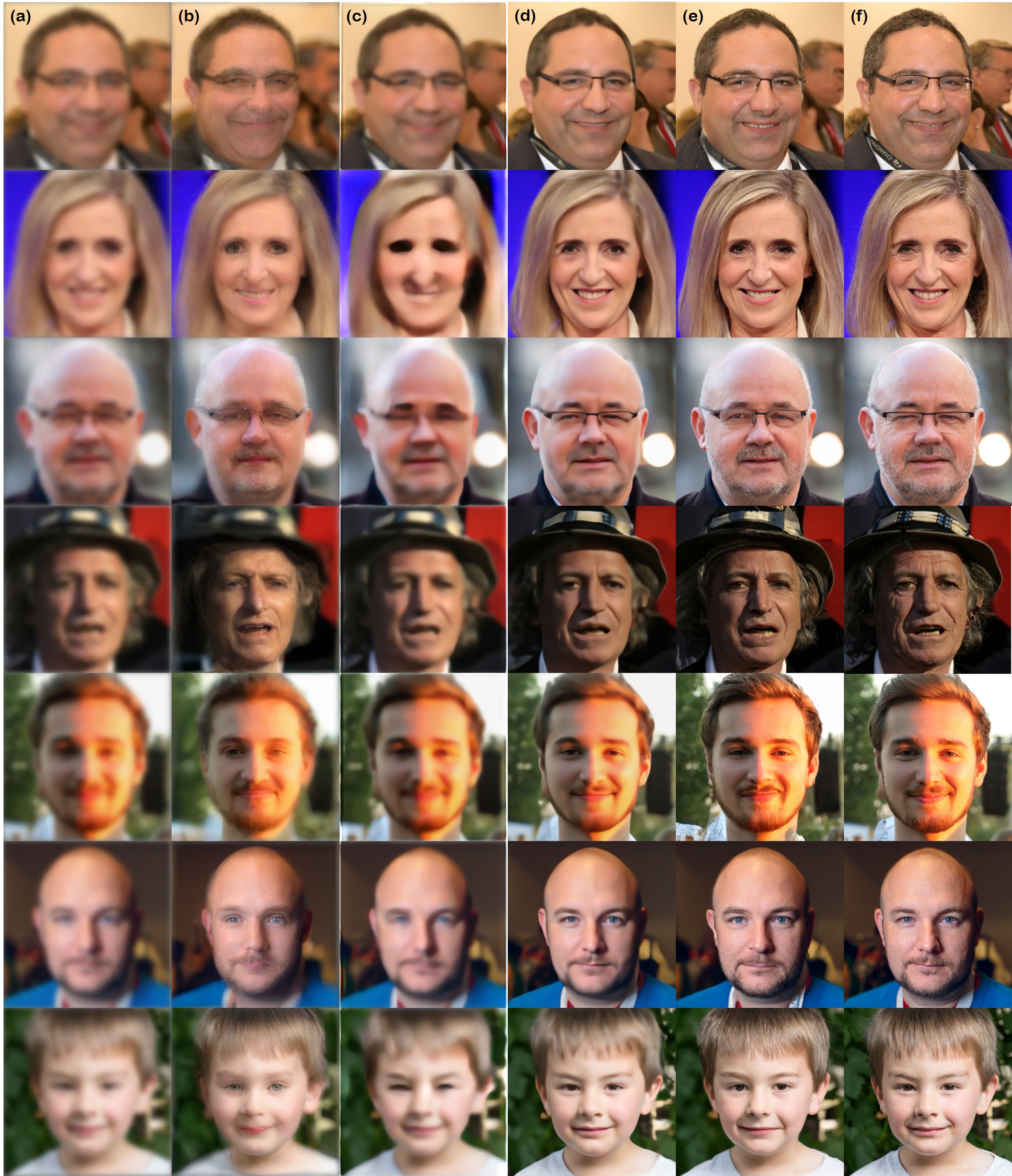}}}
  \caption{Imaging through turbulence results on the FFHQ $256\times 256$ dataset. (a) Measurement, (b) ILVR~\cite{choi2021ilvr}, (c) MPRNet~\cite{zamir2021multi}, (d) TSR-WGAN~\cite{jin2021neutralizing}, (e) BlindDPS (\textbf{ours}), (f) Ground truth.}
  \label{fig:results_turbulence_ffhq_extensive}
\end{figure*}

\begin{figure*}[t]
  \centering
    \centerline{{\includegraphics[width=1.0\linewidth]{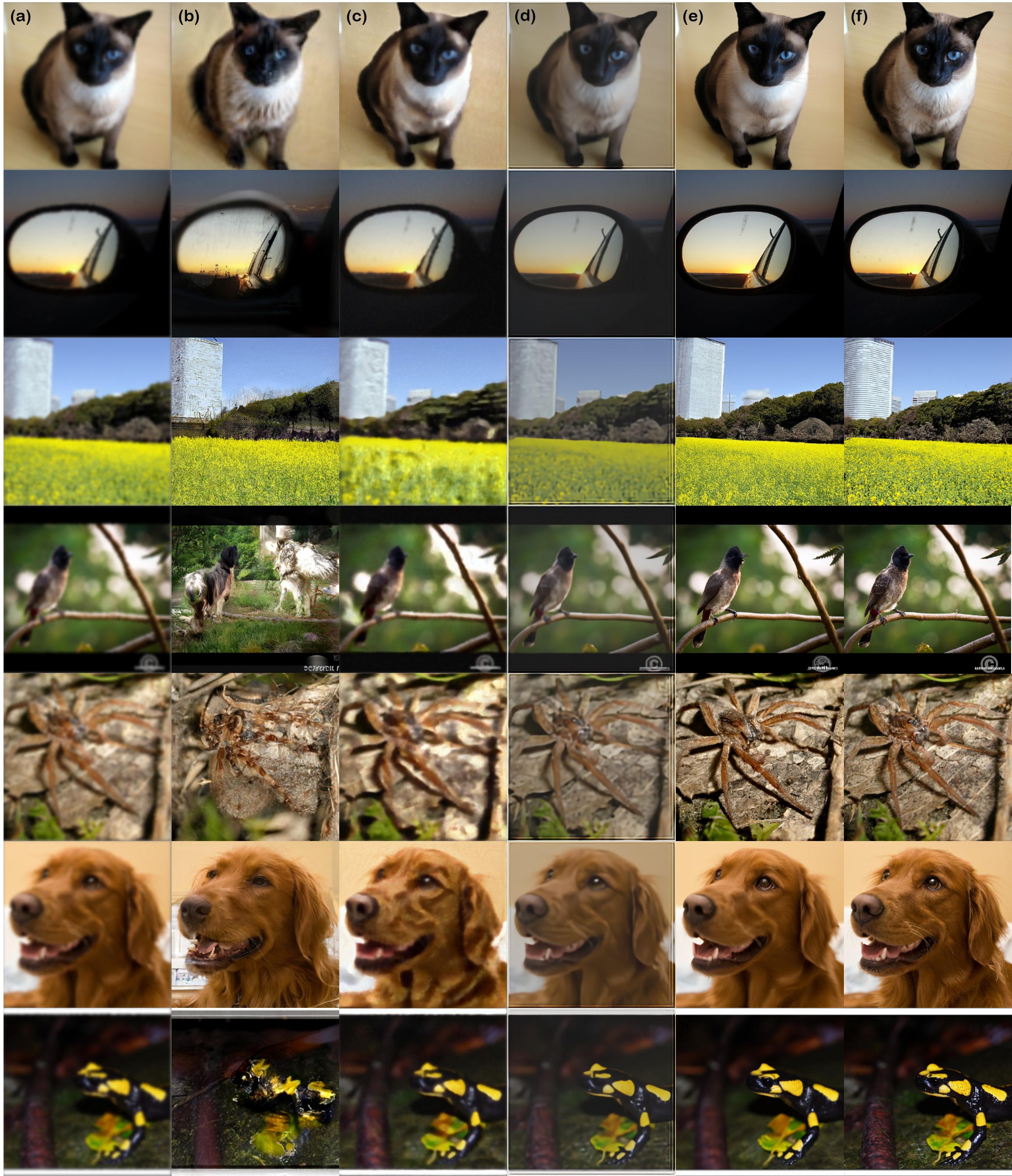}}}
  \caption{Imaging through turbulence results on the ImageNet $256\times 256$ dataset. (a) Measurement, (b) ILVR~\cite{choi2021ilvr}, (c) MPRNet~\cite{zamir2021multi}, (d) TSR-WGAN~\cite{jin2021neutralizing}, (e) BlindDPS (\textbf{ours}), (f) Ground truth.}
  \label{fig:results_turbulence_imagenet_extensive}
\end{figure*}

Further experimental results on blind deblurring are shown in Fig.~\ref{fig:results_motion_deblur_ffhq_extensive},~\ref{fig:results_motion_deblur_afhq_extensive},~\ref{fig:results_gaussian_deblur_ffhq_extensive},~\ref{fig:results_gaussian_deblur_afhq_extensive}. Further experimental results on imaging through turbulence are shown in Fig.~\ref{fig:results_turbulence_ffhq_extensive},~\ref{fig:results_turbulence_imagenet_extensive}.

\end{document}